\documentclass[Afour,sageh,times]{sagej}

\usepackage{moreverb}

\usepackage{amsmath,amssymb,subfig,url,mathrsfs,array,graphicx,bm,balance}
\usepackage[T1]{fontenc}

\newcommand{\figref}[1]{Fig.~\ref{#1}}
\newcommand{\secref}[1]{Sec.~\ref{#1}}
\newcommand{\tabref}[1]{Table~\ref{#1}}
\newtheorem{definition}{\textbf{Definition}}

\newtheorem{theorem}{\textbf{Theorem}}

\usepackage[colorlinks,bookmarksopen,bookmarksnumbered,citecolor=red,urlcolor=red]{hyperref}

\newcommand\BibTeX{{\rmfamily B\kern-.05em \textsc{i\kern-.025em b}\kern-.08em
T\kern-.1667em\lower.7ex\hbox{E}\kern-.125emX}}

\begin{document}

\runninghead{Sombolestan and Nguyen}

\title{Hierarchical Adaptive Motion Planning with Nonlinear Model Predictive Control for Safety-Critical Collaborative Loco-Manipulation}

\author{Mohsen Sombolestan\affilnum{1} and Quan Nguyen\affilnum{1}}

\affiliation{\affilnum{1}Department of Aerospace and Mechanical Engineering, University of Southern California, Los Angeles, CA}

\corrauth{Mohsen Sombolestan, Department of Aerospace and Mechanical Engineering, University of Southern California, Los Angeles, CA, 90089}

\email{somboles@usc.edu}

\begin{abstract}
As legged robots take on roles in industrial and autonomous construction, collaborative loco-manipulation is crucial for handling large and heavy objects that exceed the capabilities of a single robot. However, ensuring the safety of these multi-robot tasks is essential to prevent accidents and guarantee reliable operation. This paper presents a hierarchical control system for object manipulation using a team of quadrupedal robots. The combination of the motion planner and the decentralized locomotion controller in a hierarchical structure enables safe, adaptive planning for teams in complex scenarios. A high-level nonlinear model predictive control planner generates collision-free paths by incorporating control barrier functions, accounting for static and dynamic obstacles. This process involves calculating contact points and forces while adapting to unknown objects and terrain properties. The decentralized loco-manipulation controller then ensures each robot maintains stable locomotion and manipulation based on the planner’s guidance. The effectiveness of our method is carefully examined in simulations under various conditions and validated in real-life setups with robot hardware. By modifying the object's configuration, the robot team can maneuver unknown objects through an environment containing both static and dynamic obstacles. We have made our code publicly available in an open-source repository at \href{https://github.com/DRCL-USC/collaborative_loco_manipulation}{https://github.com/DRCL-USC/collaborative\_loco\_manipulation}.
\end{abstract}

\keywords{Safety-critical Motion Planning, Legged Robots, Collaborative Manipulation, Nonlinear MPC}

\maketitle

\section{Introduction} \label{sec: introduction}

As robots are deployed in diverse applications such as industrial facilities and autonomous construction, there will be instances where manipulating objects surpasses the actuation capabilities of a single robot \cite{Khatib1999RobotsCapabilities}. When humans face similar challenges, they collaborate to efficiently and reliably move objects too heavy for an individual to handle alone. Multi-agent systems utilizing wheeled robots are the most prevalent method for transporting payloads across well-structured terrain due to their ease of control, efficiency, and adaptable wheel configurations \cite{Tallamraju2019EnergyValidation}.
In real-world applications, collaborative loco-manipulation of large and bulky objects often involves complex maneuvers, increasing the risk of collisions with the environment. When operating near humans, it becomes even more critical to implement safety measures to prevent collisions.
We aim to enable legged robot teams to achieve similar collaborative capabilities.
\begin{figure}[t!]
	\center
	\includegraphics[width=1.0\linewidth]{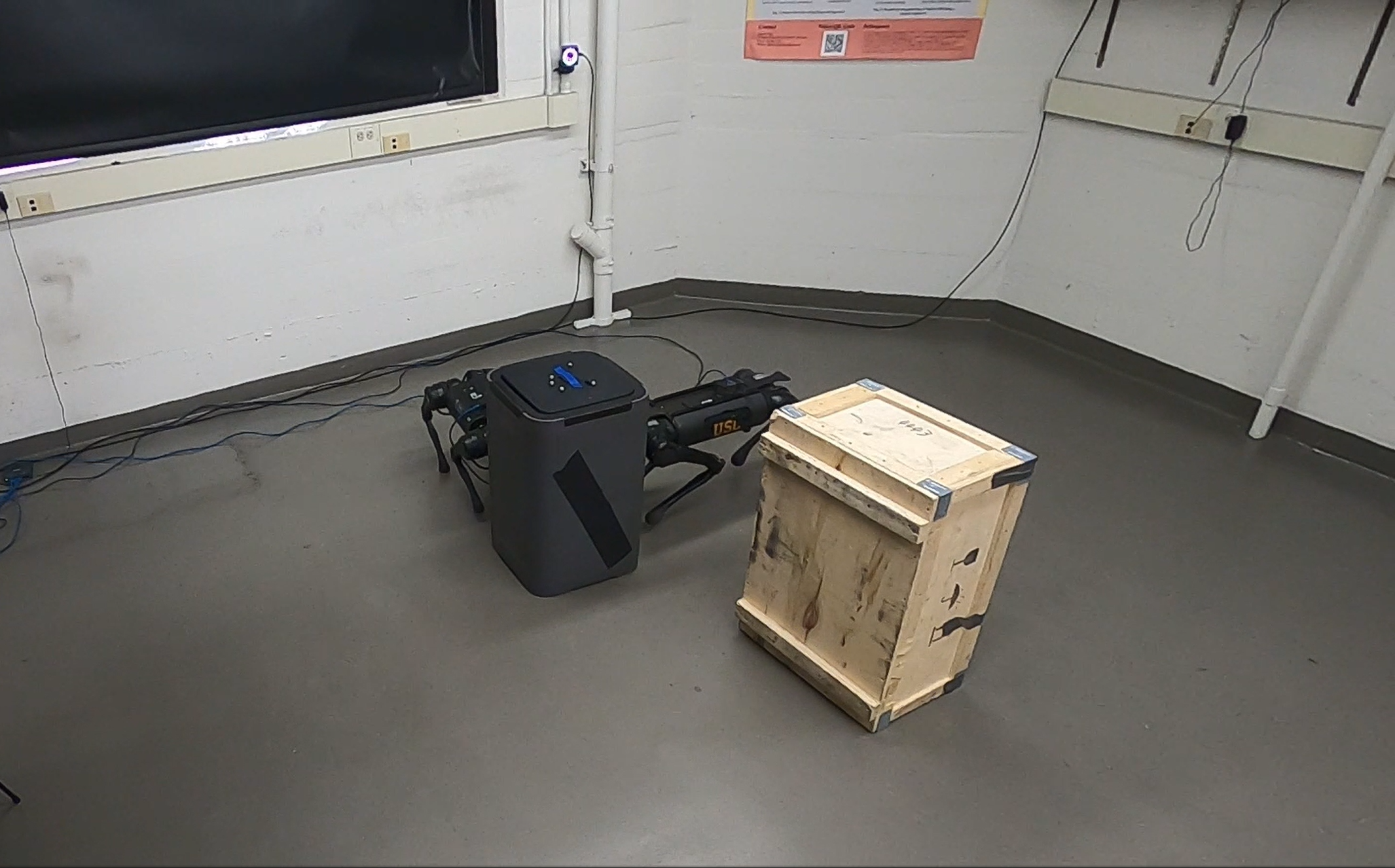}
	\caption{\textbf{Snapshots of collaborative object manipulation with safety considerations. More results presented in:} \href{https://youtu.be/cU_qevkW86I}{https://youtu.be/cU\_qevkW86I}}
	\label{fig: experiment_overview}
\end{figure}

Collaborative object manipulation has been a subject of research since the early developments in robot manipulators \cite{Tarn1986COORDINATEDARMS, Khatib1988ObjectSystem} and mobile robots \cite{Khatib1996}. Some initial studies also utilized adaptive control for collaborative manipulation in mobile robots, operating without assumptions about the object's mass through the use of a centralized controller \cite{Hu1995, Li2008}. Over time, methods evolved to employ decentralized controllers \cite{Liu1998DecentralizedCooperations, Fink2008, Verginis2017, Culbertson2021}. These techniques typically depend on a rigid connection between objects and robots during manipulation tasks. Additionally, in certain scenarios, it is necessary to measure the manipulators' relative positions from the center of mass (COM) of the object \cite{Prattichizzo2008}. 

Legged robots are recognized for their rapid movement and ease of maneuvering due to their flexible locomotion capabilities. The progress in model predictive control (MPC) for legged robots \cite{DiCarlo2018, Li2021} has enabled the development of real-time control systems capable of executing various walking gaits. Most studies on quadruped robots have focused on locomotion \cite{Focchi2017, Bledt2018} and loco-manipulation by individual robots \cite{Chiu2022, Sleiman2021, Zimmermann2021, Rigo2023, Wolfslag2020OptimisationRobots, Ferrolho2023RoLoMa:Arms}. 
In our previous works, we also addressed issues with significant uncertainties in the robot model \cite{Sombolestan2024, Sombolestan2021}, handling objects with unknown properties \cite{Sombolestan2023b}, and in this work, we aim to extend it for collaborative manipulation by a team of robots.

However, there is limited research on the collaboration among multiple quadruped robots. In scenarios where multiple general-purpose robots are available rather than specialized, larger robots, the collaboration among several quadruped robots can be highly beneficial. These robots can work together to perform tasks beyond the capabilities of a single robot, such as object manipulation in industrial factory locations and last-mile delivery operations. Research has explored using multiple quadruped robots to tow a load with cables toward a target while avoiding obstacles \cite{Yang2022b}. A recent trend involves using interconnected legged robots for collaborative manipulation with holonomically constant properties defining the configuration setup. For instance, \cite{Kim2023LayeredApproaches} developed both centralized and distributed MPCs as high-level planners, followed by a distributed whole-body tracking controller that makes two quadruped robots to carry a payload. Another study introduced a passive arm concept to facilitate collaboration between robots and between robots and humans for payload transportation \cite{Turrisi2024PACC:Control}. This research area has even expanded to bipedal robots, where researchers designed a decentralized controller using reinforcement learning for multi-biped robot carriers, adaptable to varying numbers of robots \cite{Pandit2024LearningTransport}.
However, many manipulation tasks, including those mentioned, often require prior knowledge of the manipulated object's properties and the terrain, such as mass, geometry, and friction coefficient. 

In our previous work \cite{Sombolestan2023}, a preliminary version of this research was presented, where we introduced collaborative manipulation with a team of quadrupedal robots using quadratic programming (QP)-based force distribution but without incorporating safety considerations. In this work, which builds upon our previous work, we address the challenge of manipulating objects with uncertain characteristics to navigate through an environment filled with dynamic and static obstacles to ensure safety and stability throughout the manipulation process. Ensuring safety requires optimizing the path of the manipulated object to avoid collisions with obstacles, as well as preventing collisions between each robot and the obstacles. Our method is scalable to accommodate any number of robots, providing flexibility in operation. Some snapshots demonstrating our results of manipulating an object using two robots in an environment with an obstacle are shown in \figref{fig: experiment_overview}.

Our method comprises a high-level adaptive motion planner employing Model Predictive Control (MPC) and a loco-manipulation tracking controller. The planner addresses real-world challenges in object manipulation, particularly regarding safety and obstacle avoidance. The safety aspect is guaranteed through control barrier functions integrated into the MPC planner, allowing it to avoid collisions with dynamic obstacles. The system considers not only the potential collisions between the manipulated object and obstacles but also ensures each robot avoids obstacles independently.

Moreover, our approach does not rely on assumptions about object properties such as mass, inertia, and the center of mass (COM) location, nor on terrain properties like the friction coefficient. These factors are managed through our adaptive dynamic formulation, derived from an adaptive law, and stability is guaranteed via a control Lyapunov function. This comprehensive motion planning provides optimal manipulation forces and contact point locations for each robot, enabling safe maneuvering of the manipulated object in the environment with obstacles to reach the target location.

The remainder of the paper is organized as follows. First, in \secref{sec: problem_statement}, we define our problem statement explicitly, outlining all assumptions and requirements. Next, in \secref{sec: Preliminaries}, we provide the necessary preliminaries to ensure the paper is accessible. Following this, we present an overview of our control system in \secref{sec: overview}. In \secref{sec: motion planner}, we explain the component of our proposed motion planner, and in \secref{sec: NMPC framework}, we elaborate on the formulation of the nonlinear model predictive control for the motion planner. We then present our results, achieved both in simulation and on robot hardware, in \secref{sec: results}. Finally, concluding remarks are provided in \secref{sec: conclusion}.
\section{Problem Statement} \label{sec: problem_statement}

\begin{figure}[t!]
	\center	\includegraphics[width=1.0\linewidth]{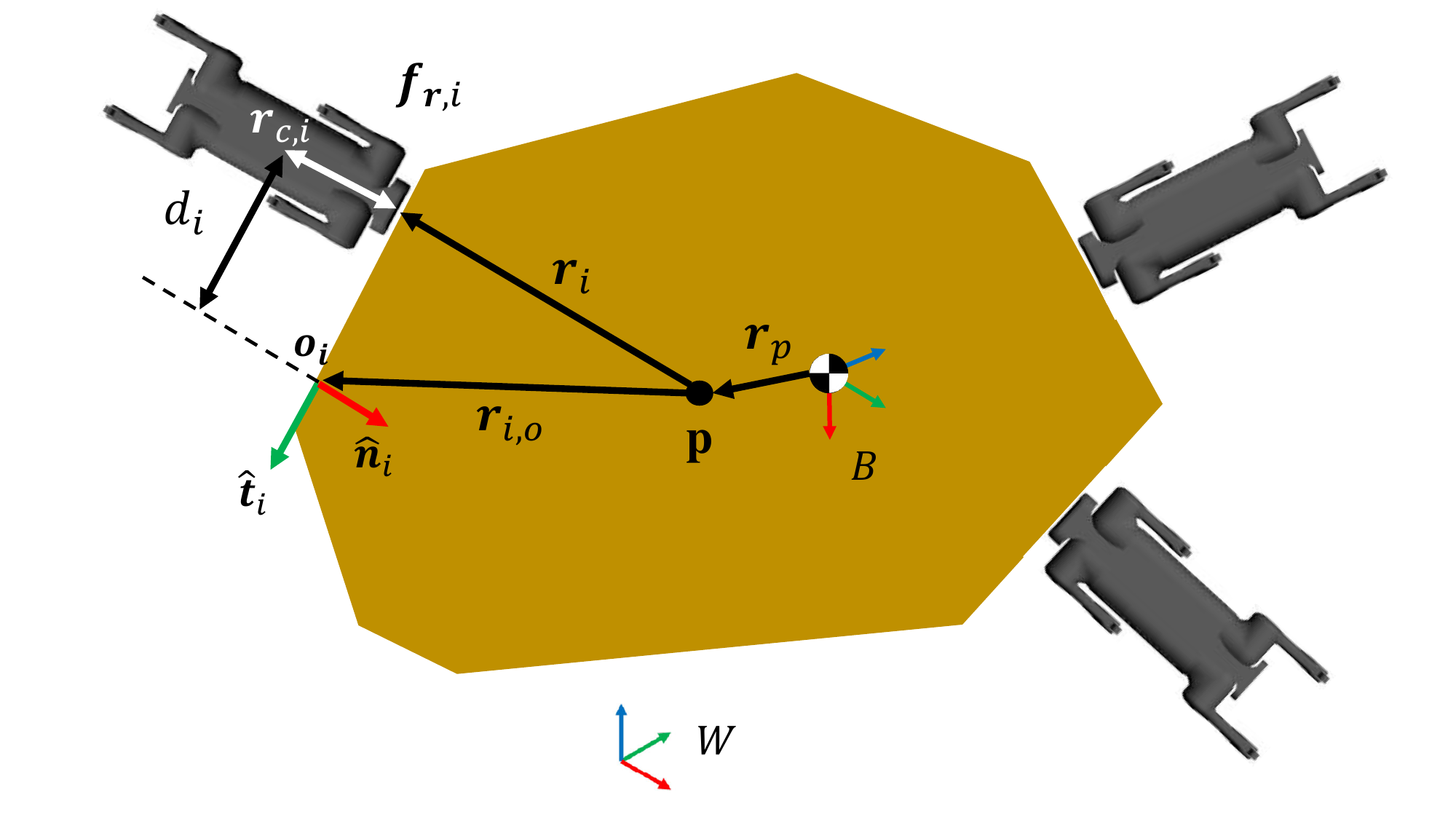}
	\caption{\textbf{Schematic of collaborative object manipulation}}
	\label{fig: object_schematic}
\end{figure}

We aim to manipulate a rigid object with unspecified inertia characteristics, including mass ($m_b$), body-frame inertia about the center of mass (COM) ($\bm{I}_G$), and the location of the COM ($\bm{r}_p$). Our objective is to control this object using a group of $N_r$ agents, each indexed by $i\in \{1, \ldots, N_r\}$. A visual representation of the object with multiple agents is depicted in \figref{fig: object_schematic}. The number of robots involved can vary, as our method allows for the distribution of force among any number of robots. 

In this scenario, we assume the object possesses a flat surface suitable for robot contact. Two frames, the world frame $W$ and the body frame $B$, are associated with the object's COM. A body-fixed point $\bm{p}$ serves as the reference for all measurements related to the object and can be arbitrarily chosen. Additionally, we have estimates of each corner's location on the object, defining the allowable range of motion for each robot along the object's surface to ensure they do not surpass its boundaries. Moreover, each robot is capable of providing its own state estimation, allowing us to determine the contact point location $\bm{r}_{i}$ relative to the reference point $\bm{p}$. The robots have freedom of movement along the object's surface, indicating they are not rigidly attached. Furthermore, we assume negligible friction between the robots and the object at the contact points. Therefore, each robot can only apply a perpendicular force $\bm{f}_{r,i}$ onto the object's surface (along $\hat{\bm{n}}_{i}$) while undergoing tangential movement $d_i$ along the object surface (along $\hat{\bm{t}}_{i}$). 
We consider the initial contact point from which each robot starts as the origin of the contact frame ($\bm{o}_i$), and the distance $d_i$ is measured from this point. This initial location can be represented as $\bm{r}_{i,o}$ with respect to the reference point $p$.
Note that all vectors mentioned in \figref{fig: object_schematic}, such as $\bm{f}_{r,i}$, $\bm{r}_{0,i}$, $\bm{r}_p$, $\hat{\bm{n}}_{i}$, and $\hat{\bm{t}}_{i}$, are represented in the body frame $B$ and they are two-dimensional vectors.

\begin{figure}[t!]
	\center	\includegraphics[width=1.0\linewidth]{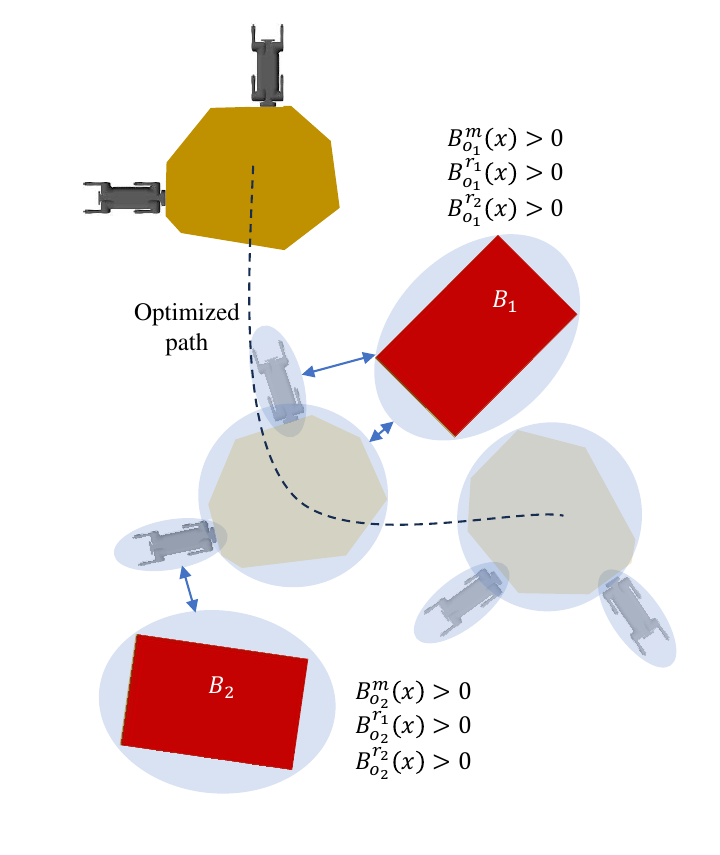}
	\caption{\textbf{Schematic of object manipulation considering the safety} Each barrier function $B$ specifies the safety requirement for agents or manipulated objects with respect to the obstacles}
	\label{fig: safety_schematic}
\end{figure}

Now, let us consider the scenario where the robot navigates through an environment containing $N_o$ obstacles, denoted by $j\in \{1, \ldots, N_o\}$. The schematic illustrating multiple agents within this environment alongside multiple obstacles is presented in \figref{fig: safety_schematic}. Ensuring safety involves optimizing the path of the manipulated object to prevent any collisions between that and the obstacles. Additionally, we must account for the potential collision of each agent with the obstacles. Therefore, for each obstacle, it is necessary to define $N_r + 1$ barrier functions, totaling $N_o(N_r + 1)$, to ensure comprehensive non-collision and safety for the entire system. The specifics of each barrier function's definition will be elaborated upon in \secref{sec: CBF Constraints}.

\section{Preliminaries} \label{sec: Preliminaries}

\subsection{Equation of Motion for a Rigid Object}
The equation motion of a rigid body object can be written as follows:
\begin{subequations}
\begin{align}
&\bm{F} = m_b {\bm{\ddot{x}}}_G
\\
&\bm{M}_G = \bm{R} \bm{I}_G \bm{R}^T \dot{\bm{\omega}} + \bm{\omega} \times (\bm{R} \bm{I}_G \bm{R}^T \bm{\omega})
\end{align}    
\end{subequations}
where $\bm{R}$ is the rotation matrix from body frame $B$ to the world frame $W$, $\bm{\omega}$ is the angular velocity of the object, and $\dot{\bm{\omega}}$ is the angular acceleration. Since, in our problem, the COM position is unknown, we should derive the equation of motion of the rigid object with respect to the reference point $\bm{p}$:
\begin{subequations}
\begin{align}
\label{eq: translation}
&\bm{F} = m_b \bm{\ddot{x}}_p - m_b(\dot{\bm{\omega}} \times \bm{R}\bm{r}_p) - m_b \bm{\omega} \times (\bm{\omega} \times \bm{R} \bm{r}_p) 
\\
 \label{eq: rotational}
&\bm{M}_p = \bm{R} \bm{I}_p \bm{R}^T \dot{\bm{\omega}} + \bm{\omega} \times (\bm{R} \bm{I}_p \bm{R}^T \bm{\omega}) - m_b \bm{R} \bm{r}_p \times \bm{\ddot{x}}_p
\end{align}
\end{subequations}
where $\bm{F}$ and $\bm{M}_p$ are the force and moment required for object manipulation, respectively, $\ddot{\bm{x}}_p$ is the object's linear acceleration at point $\bm{p}$, and $\bm{I}_p$ is the object's moment of inertia with respect to $\bm{p}$.

Considering that the object is on the ground and will be manipulated within planar coordinates, we can limit our focus to the planar aspect of the equation of motion. To achieve this, we define the configuration variable $\bm{q}_b = [\bm{x}_p, \theta]$, where $\bm{x}_p$ is the position of reference point $\bm{p}$ in the world frame and $\theta$ represents the object's yaw angle. We also take into account an external wrench $\bm{f}_k$ and express the equation of motion in a compact form as follows:
 \begin{equation}
 \label{eq: equation of motion}
\bm{\tau} = \bm{H}(\bm{q}_b)\ddot{\bm{q}_b} + \bm{C}(\bm{q}_b, \dot{\bm{q}_b})\dot{\bm{q}_b}  + \bm{f}_k
\end{equation}
where $\bm{\tau}$ is the wrench applied to the rigid object from a team of robots. 

The control barrier functions can benefit from the expression of system dynamics in a control-affine format. Considering $\bm{x}_b = [\bm{q}_b, \dot{\bm{q}}_b]^T$, then, we have a control-affine system as follow:
 \begin{equation}
 \label{eq: dynamic_equation_control_affine}
 \dot{\bm{x}}_b = f_b(\bm{x}_b)  +  g_b(\bm{x}_b) \bm{\tau} 
\end{equation}
where:
\begin{subequations}
\begin{align}
 f_b(\bm{x}_b) &= 
 \left[ \begin{array}{c} \dot{\bm{q}}_b \\  - \bm{H}(\bm{q}_b)^{-1}(\bm{C}(\bm{q}_b, \dot{\bm{q}_b})\dot{\bm{q}_b}  + \bm{f}_k)  \end{array} \right]
  \\
 g_b(\bm{x}_b) &= \left[ \begin{array}{c} \bm{0}_{3 \times 3} \\  \bm{H}(\bm{q}_b)^{-1}  \end{array} \right]
\end{align}
\end{subequations}

\subsection{Control Barrier Functions (CBFs)}

Control Barrier Functions (CBFs) serve as a mechanism for generating controllers that are demonstrably safe, akin to the manner in which Control Lyapunov Functions (CLFs) \cite{Sontag1999} ensure stability. In various respects, CBFs mirror the extension of Lyapunov functions to CLFs. We consider the system \eqref{eq: dynamic_equation_control_affine} safe if $\bm{x}_b$ remains within a safe set $\mathcal{S}$ for all time. The theoretical definition for this criterion is that $\mathcal{S}$ is \textit{forward invariant}, signifying that for any initial state $\bm{x}_b(0)$ within $\mathcal{S}$, $\bm{x}_b(t)$ stays within $\mathcal{S}$ for all $t \geq 0$ \cite{Ames2014ControlControl}.

The determination of the safe set relies on the system configuration, which, in our scenario, includes the manipulated object and the positions of each robot to avoid collisions with nearby obstacles. Therefore, the safe set $\mathcal{S}$ can be defined as a continuously differentiable function denoted by $B(\bm{x}_b)$:
\begin{subequations}\label{eq: safe_set}
\begin{align} 
    \mathcal{S} := & \{ \bm{x}_b : B(\bm{x}_b) \geq 0 \}, \\
    \partial \mathcal{S} := & \{ \bm{x}_b : B(\bm{x}_b) = 0 \}. 
\end{align}
\end{subequations}

\begin{definition} \label{def:cbf} 
A continuously differentiable function $B$ is a \textit{control barrier function (CBF)} for \eqref{eq: dynamic_equation_control_affine} if there exists an extended class $\mathcal{K}_\infty$ function $\alpha$ such that \cite{Ames2017ControlSystems}:
    \begin{equation}\label{eq: cbf_definition}
        \sup_{\tau} \,\,\,\, \dot{B}(\bm{x}_b,\bm{\tau}) \geq -\alpha(B(\bm{x}_b)),
    \end{equation}
    for all $\bm{x}_b \in \mathcal{S}$, where 
    \begin{equation}
    \dot{B}(\bm{x}_b,\bm{\tau}) = \underbrace{\frac{\partial B(\bm{x}_b)}{\partial \bm{x}_b}f_b(\bm{x}_b)}_{:= L_f B(\bm{x}_b)} + 
    \underbrace{\frac{\partial B(\bm{x}_b)}{\partial \bm{x}_b}g_b(\bm{x}_b)}_{:=L_g B(\bm{x}_b)} \bm{\tau}
    \end{equation}
    with $L_f B(\bm{x}_b)$ and $L_g B(\bm{x}_b)$ the Lie derivatives \cite{Slotine1991} of $B$ with respect to $f_b(\bm{x}_b)$ and $g_b(\bm{x}_b)$, respectively. 
\end{definition}

The function $B$ as defined in equation \eqref{eq: safe_set} corresponds specifically to a Zeroing Control Barrier Function (ZCBF). Another type of barrier function known as Reciprocal Control Barrier Function (RCBF), as described by \cite{Ames2017ControlSystems}, tends to become unbounded at the boundaries of the set, meaning that $B(\bm{x}_b)$ approaches infinity as $\bm{x}_b$ approaches the boundary $\partial \mathcal{S}$. Research has shown that ZCBFs offer simpler and smoother performance compared to RCBFs in real-world systems.

\begin{theorem}\label{th: cbf}
If $B$ is a \textit{CBF} for \eqref{eq: dynamic_equation_control_affine}, then any locally Lipschitz continuous controller $\bm{\tau}=k(\bm{x}_b)$ satisfying
\begin{equation} \label{eq: cbf_safe_condition}
    \dot{B}(\bm{x}_b, k(\bm{x}_b)) \geq -\alpha(B(\bm{x}_b)), \qquad \forall \bm{x}_b\in \mathcal{S}
\end{equation}
guarantees that \eqref{eq: dynamic_equation_control_affine} is safe with respect to $\mathcal{S}$ \cite{Ames2017ControlSystems}.
\end{theorem}

Now, let us consider the function \( B(\bm{x}_b) \) to have an arbitrarily high relative degree \( r \geq 1 \), which implies
\begin{equation}
    B^{(r)}(\bm{x}_b, \bm{\tau}) = L_f^r B(\bm{x}_b) + L_g L_f^{r-1} B(\bm{x}_b) \bm{\tau}
\end{equation}
where $L_g L_f^{r-1} B(\bm{x}_b) \neq 0$ and $L_g L_f B = L_g L_f^2 B = \dots = L_g L_f^{(r-2)} B = 0 $. Let us define:
\begin{equation}
    \bm{\eta}(\bm{x}_b) := \left[ \begin{array}{c}
    B(\bm{x}_b) \\
    \dot{B}(\bm{x}_b) \\
    \ddot{B}(\bm{x}_b) \\
    \vdots \\
    B^{(r-1)}(\bm{x}_b)
    \end{array} \right] = \left[ \begin{array}{c}
    B(\bm{x}_b) \\
    L_f B(\bm{x}_b) \\
    L_f^2 B(\bm{x}_b) \\
    \vdots \\
    L_f^{r-1} B(\bm{x}_b)
    \end{array} \right]
\end{equation}
Then, by defining \( \bm{\mu} = L_f^r B(\bm{x}_b) + L_g L_f^{r-1} B(\bm{x}_b) \bm{\tau} \), we obtain:
\begin{equation}
\label{fgBarDynamics}
\dot{\bm{\eta}} = \bm{D}_b \bm{\eta} + \bm{G}_b \bm{\mu}, \nonumber \\
B(\bm{x}_b) = \bm{C}_b \bm{\eta},
\end{equation}
where
\begin{eqnarray}
    \bm{D}_b &=& \begin{bmatrix}
        0 & 1 & 0 & \cdots & 0 \\
        \vdots & \vdots & \vdots & \ddots & \vdots \\
        0 & 0 & 0 & \cdots & 1 \\
        0 & 0 & 0 & \cdots & 0 \\
    \end{bmatrix}, \quad
    \bm{G}_b = \begin{bmatrix}
        0 \\
        \vdots \\
        0 \\
        1
    \end{bmatrix}, \\
    \bm{C}_b &=& \begin{bmatrix} 1 & 0 & \cdots & 0 \end{bmatrix}. \nonumber
\end{eqnarray}
By choosing \( \bm{\mu} \geq - \bm{K}_\alpha \bm{\eta}(\bm{x}_b) \), it follows that \( B(\bm{x}_b) \geq \bm{C}_b e^{(\bm{D}_b - \bm{G}_b \bm{K}_\alpha)t} \bm{\eta}(\bm{x}_{b,0}) \). These derivations lead to the following definition \cite{Ames2019ControlApplications}.

\begin{definition} \label{def:ecbf}
 A $r$-times continuously differentiable function $B$ is an \textit{exponential control barrier function (ECBF)} for \eqref{eq: dynamic_equation_control_affine} if there exists a row vector $\bm{K}_\alpha \in \mathbb{R}^r$ such that $\forall \bm{x}_b \in \{ \mathcal{S} \setminus  \partial \mathcal{S} \}$:
\begin{equation}
	\label{eq:ecbf_definition}
	\sup_{\bm{\tau}}  \left[ L_f^r B(\bm{x}_b) + L_g L_f^{r-1} B(\bm{x}_b) \bm{\tau} \right] \geq - \bm{K}_\alpha \bm{\eta}(\bm{x}_b)
\end{equation} where results in $B(\bm{x}_b) \ge \bm{C}_b e^{(\bm{D}_b-\bm{G}_b \bm{K}_\alpha)t} \bm{\eta}(\bm{x}_{b,0})$ whenever $B(\bm{x}_{b,0}) \ge 0$.
\end{definition}

Note that for $r=1$, the definition \ref{def:ecbf} becomes the same as definition \ref{def:cbf} with a simple and specific instance of $\alpha(B(\bm{x}_b))$ is $\alpha B(\bm{x}_b)$, where $\alpha > 0$.

\begin{figure*}[t!]
	\center \includegraphics[width=1\linewidth]{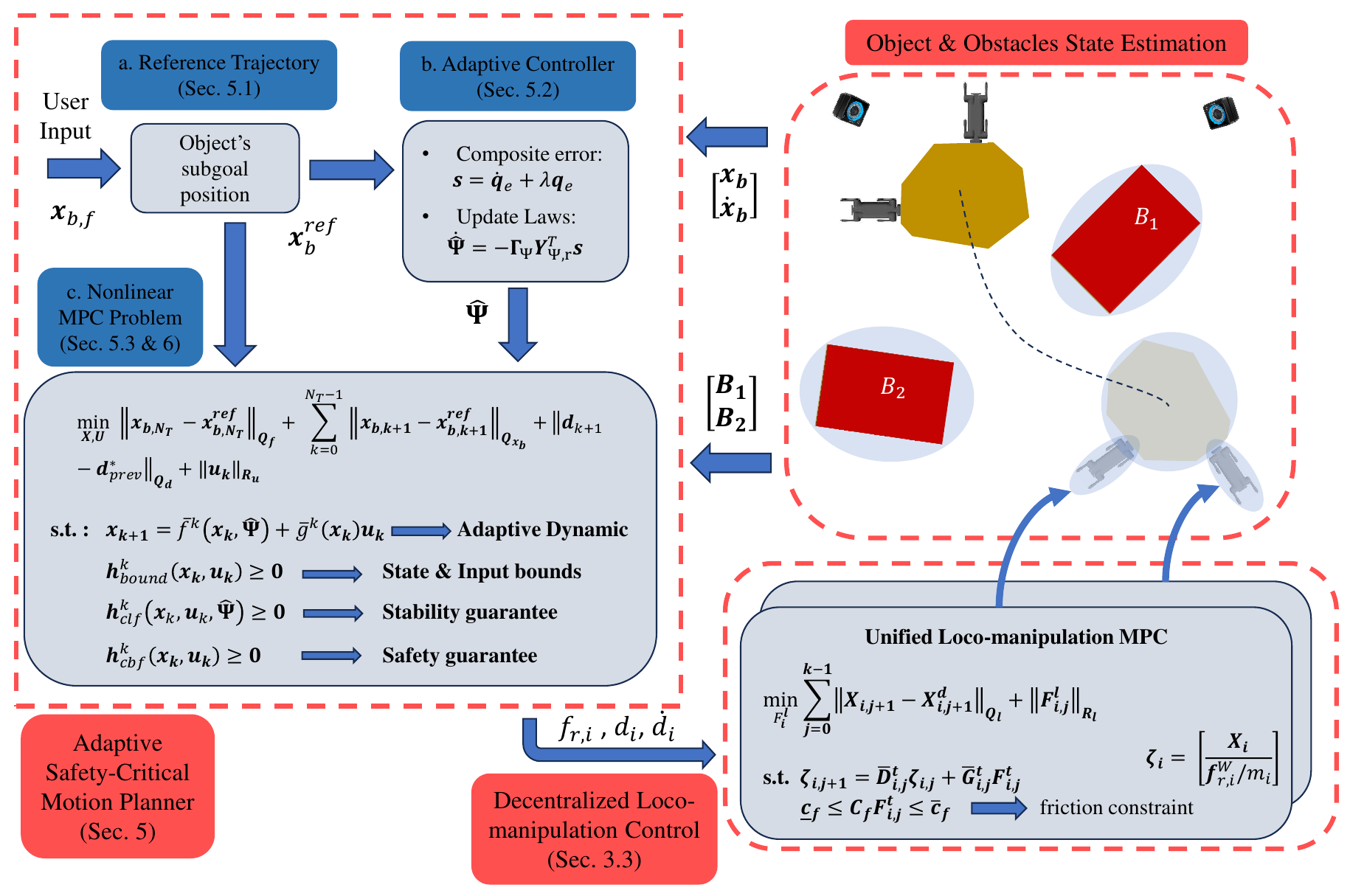}
	\caption{{\bfseries{Block diagram of our proposed approach}}. Our approach includes 1) state estimation for object and obstacle states; 2) a safety-critical motion planner that utilizes an adaptive controller, CLF, and CBFs for team safety and obstacle avoidance within an MPC framework; and 3) a decentralized loco-manipulation controller that employs a unified MPC for simultaneous stable locomotion and manipulation.}
	\label{fig: block_diagram}
\end{figure*}

\subsection{Decentralized Loco-manipulation Control} \label{sec: loco-manipulation control}
The decentralized loco-manipulation control method developed in \cite{Sombolestan2023b} will serve as the controller for locomotion and manipulation for each agent. This method employs a unified MPC responsible for regulating the manipulation force while ensuring the robot's balance.

The equation of motion for each robot, incorporating manipulation forces, is derived based on the state representation introduced in \cite{DiCarlo2018}:
\begin{equation}\label{eq: loco-manipulation dynamic}
 \bm{\dot{X}}_i = \bm{D}_i \bm{X}_i + \bm{G}_i \bm{F}_{i}^l + \bm{f}_{r,i}^W / m_i
\end{equation}
Here $i\in \{1, \ldots, N_r\}$ is the robot's index, $\bm{f}_{r,i}^W$ represents the manipulation force vector in the world frame, $m_i$ denotes the robot mass, $\bm{F}_{i}^l$ denotes the vector of ground reaction forces for all the legs, and $\bm{X}_i$ contains the robot's body's COM location, Euler angles, and velocities. Further details regarding this equation and the definitions of $\bm{D}$ and $\bm{G}$ are provided in \cite{Sombolestan2023b}.

The MPC technique employs linear discrete-time dynamics to predict the system's behavior over a finite time horizon. However, utilizing a conventional discretization method such as zero-order hold necessitates integrating the manipulation term $\bm{f}_{r,i}^W$ from equation \eqref{eq: loco-manipulation dynamic} into the state vector, thereby extending it for MPC formulation. Thus, equation \eqref{eq: loco-manipulation dynamic} can be reformulated as:
\begin{equation}
\label{eq: combined SS}
\bm{\dot{\zeta}}_i = \bar{\bm{D}}_i \bm{\zeta}_i + \bar{\bm{G}}_i \bm{F}_{i}^l 
\end{equation}
where
\begin{subequations}
\begin{align} \label{eq: extended SS components}
&\bm{\zeta}_i = \left[\begin{array}{c} 
    \bm{X}_i \\ 
    \hline
    \bm{f}_{r,i}^W /m_i
    \end{array} \right] \in \mathbb{R}^{15}
\\
&\bar{\bm{D}}_i = \left[\begin{array}{@{}c|c@{}} 
        \begin{matrix}
           \bm{D}_{i} \in \mathbb{R}^{13 \times 13}
        \end{matrix}
        & \begin{matrix}
            \bm{0}_{6 \times 2} \\
            \bm{I}_{2 \times 2} \\
            \bm{0}_{5 \times 2} \\
        \end{matrix}
    \\
    \hline
    \bm{0}_{2 \times 13} & \bm{0}_{2 \times 2}
\end{array} \right] \in \mathbb{R}^{15 \times 15}
\\ 
&\bar{\bm{G}}_i = \left[\begin{array}{c}
    \bm{G}_i \\
    \hline
    \bm{0}_{2 \times 12}
\end{array} \right] \in \mathbb{R}^{15 \times 12}
\end{align}
\end{subequations}
and $\bm{\zeta}_i$ represents the augmented vector. Therefore, a linear MPC can be formulated as follows:
\begin{subequations}
\begin{align}
\label{eq: mpc}
\min_{\bm{F}_i^{l}} \quad & \sum_{j=0}^{k-1} 
\|\bm{X}_{i,j+1} - \bm{X}_{i,j+1}^d\|_{\bm{Q}_{l}} +
\|\bm{F}_{i,j}^l\|_{\bm{R}_{l}} \\ 
\textrm{s.t.} \quad & \bm{\zeta}_{i,j+1} = \bar{\bm{D}}_{i,j}^t \bm{\zeta}_{i,j} + \bar{\bm{G}}_{i,j}^t \bm{F}_{i,j}^l \\ 
& \underline{\bm{c}}_f \leq \bm{C}_f \bm{F}_{i,j}^l \leq \bar{\bm{c}}_f
\end{align}
\end{subequations}
where $k$ denotes the number of prediction horizons, $\bm{X}_{i}^d$ represents the desired state of the robot, $\bm{Q}_{l}$ and $\bm{R}_{l}$ are diagonal positive semi-definite matrices, $\bar{\bm{D}}_{i}^t$ and $\bar{\bm{G}}_{i}^t$ denote discrete-time system dynamics matrices, and $\underline{\bm{c}}_f \leq \bm{C}_f \bm{F}_{i}^l \leq \bar{\bm{c}}_f$ represents the friction cone constraints outlined in \cite{Focchi2017}. The desired state $\bm{X}_{i}^d$ is obtained through the optimal contact point location necessary for object manipulation.

\section{System Overview: Hierarchical Adaptive Motion Planning for Safe Collaborative Object Manipulation} \label{sec: overview}

This section will briefly overview our proposed hierarchical adaptive motion planning for the collaborative manipulation of an unknown object using a team of quadruped robots. The object experiences both translational and rotational motion, guided by a team of $N_r$ robots. Our proposed method is illustrated in \figref{fig: block_diagram}. 

The control system comprises three main components as shown in \figref{fig: block_diagram}. 
The state estimator provides the object's state as the MPC's state vector for the motion planner, along with the states of obstacles needed to ensure collision avoidance within the CBF constraint. Depending on the type of obstacles, which can be either static or dynamic, the state estimator can function as either a dynamic or static estimator.

The main component of our proposed control system is the adaptive safety-critical motion planner, which generates an optimized path for the object using an MPC approach. It consists of three parts. The reference trajectory generator receives the object's goal position and orientation from the user and, based on the distance to the final goal and the object's current state, generates incremental subgoals for the MPC to ensure a smooth transition to the goal. Remember that the object's properties, such as inertia and friction, are unknown to the control system. Therefore, there is an adaptive controller that updates the uncertainty vector based on the object's deviation from the desired trajectory and incorporates this updated uncertainty vector into the dynamic equation within the MPC formulation. Finally, the object's motion planning, which includes optimizing each agent's force and contact point location, is carried out through a nonlinear MPC formulation. This optimization considers stability and safety to prevent collisions between each robot and the object with obstacles. Optimizing the contact point locations ensures collision-free movement and proper object maneuvering along the desired trajectory.
Additionally, to ensure the optimized path for the object is feasible and achievable by the team of robots, we have included interaction constraints between the robots and the object. These constraints, such as the direction and magnitude of the force as well as the torque provided by each robot, are incorporated into the dynamic equation constraints in the MPC. Since each robot is not physically connected to the object, it cannot directly apply torque to adjust the object's orientation. Instead, the required torque is generated by adjusting the force and contact point locations to maneuver the object correctly. Moreover, because the object's center of mass is unknown, the adaptive controller plays a crucial role in compensating for the uncertainty in the COM location. The details of our proposed motion planner will be elaborated in \secref{sec: motion planner} and \ref{sec: NMPC framework}.

Finally, each robot's optimized force and contact point location will be fed to a decentralized loco-manipulation controller for implementation on each robot. The structure of the loco-manipulation controller is similar to the one presented in our previous works \cite{Sombolestan2023b}. Each robot utilizes a unified MPC for locomotion, with the required manipulation force integrated into the same MPC. The advantage of this unified MPC for both locomotion and manipulation is that it regulates the manipulation force without compromising the robot's stability. The details of the loco-manipulation controller are presented in \secref{sec: loco-manipulation control}.

\section{Adaptive Safety-Critical Motion Planner} \label{sec: motion planner}

In this section, we will elaborate on our adaptive safety-critical motion planner. The motion planner receives the goal configuration for the manipulated object from the user and, based on the state measurements of the object and obstacles in the environment, optimizes the path for the object to reach the target. This optimization, formulated as a nonlinear MPC problem, considers all safety constraints, including obstacle avoidance for both the object and the team of robots, as well as constraints related to the interaction between the robots and the object. Object properties like mass and friction coefficients may be unknown or uncertain in the planner's model, so we employ an adaptive controller to compensate for these uncertainties. As shown in \figref{fig: block_diagram}, the motion planner consists of three components: the reference trajectory, the adaptive controller, and the nonlinear MPC. In the following subsections, we will elaborate on the details of each component.

\subsection{Reference Trajectory} \label{sec: Reference Trajectory}
The reference trajectory module aims to provide a smooth target trajectory based on the final goal provided by the user. Suppose we receive the goal of \(\bm{x}_{b,f}\) from the user. Typically, the target velocity is zero, so we only receive \(\bm{q}_{b,f}\), which contains \(\bm{x}_{p,f}\) and \(\theta_f\). The next step is to estimate the time required to reach the target. First, we consider feasible average linear \(v_{\text{avg}}\) and rotational \(\omega_{\text{avg}}\) velocities for the object to be manipulated. Then, we can calculate the estimated time to reach the target as follows:
\begin{center}
\begin{subequations}
  \begin{gather}
    t_{\text{avg},v} = \frac{\|\bm{x}_p - \bm{x}_{p,f}\|}{v_{\text{avg}}}, \quad
    t_{\text{avg},w} = \frac{\|\theta - \theta_f\|}{\omega_{\text{avg}}} \\
    t_{\text{avg}} = \max (t_{\text{avg},v}, t_{\text{avg},w}) 
  \end{gather}
\end{subequations}
\end{center}

Using the estimated time to reach the target $t_{\text{avg}}$ and the horizon time $T$ for the motion planner MPC, the reference trajectory module generates subgoal configurations $\bm{x}_{b}^{ref}$ for the MPC problem to make the target smoother and more feasible for the MPC to follow based on the distance from the target.

\subsection{Adaptive Control for Object Manipulation} \label{sec: adaptive controller}
In adaptive control for manipulators \cite{Slotine1991}, a linear combination of position and velocity error, denoted as \(\bm{s}\), is commonly used. This method results in exponentially stable dynamics once the surface \(\bm{s}=0\) is reached. Therefore, we define the composite error as follows:
\begin{equation} \label{eq: composite error}
\bm{s} = \bm{\dot{q}}_e + \lambda \bm{q}_e 
\end{equation}
where $\bm{q}_e$ and $\bm{\dot{q}}_e$ represented the tracking error for $\bm{q}_b$ and $\bm{\dot{q}}_b$, respectively and $\lambda$ is a positive scalar value. Then, we define the reference velocity as follows:
\begin{equation} \label{eq: reference velocity}
\bm{\dot{q}}_r = \bm{\dot{q}_b} - \bm{s}
\end{equation}

The dynamic equation \eqref{eq: equation of motion} can have model uncertainty in practice. We can separate this equation into two parts: the known nominal model and the unknown part, as follows:
\begin{align} \label{eq: uncertain_dynamic}
\bm{\tau} = &\underbrace{\bar{\bm{H}}(\bm{q}_b)\ddot{\bm{q}_b} + \bar{\bm{C}}(\bm{q}_b, \dot{\bm{q}_b})\dot{\bm{q}_b}}_{\text{nominal}} + \nonumber \\&
\underbrace{\tilde{\bm{H}}(\bm{q}_b)\ddot{\bm{q}_b} + \tilde{\bm{C}}(\bm{q}_b, \dot{\bm{q}_b})\dot{\bm{q}_b} + \bm{f}_k}_{\text{unknown}}
\end{align}
where $\bar{\bm{H}}$ and $\bar{\bm{C}}$ represent the nominal values, and $\tilde{\bm{H}}$ and $\tilde{\bm{C}}$ represent the unknown model. Note that we consider all the friction to be uncertainty within our model.

The part with uncertainty can be parameterized based on an unknown parameter vector \(\bm{\Psi}\) \cite{Slotine1991}. Thus, we can decompose the equation of motion with uncertainty into a known regressor matrix \(\bm{Y}_{\Psi, b}\) and a vector of unknown parameters \(\bm{\Psi}\):
\begin{equation} \label{eq: regressor nominal}
\tilde{\bm{H}}\ddot{\bm{q}}_b + \tilde{\bm{C}}\dot{\bm{q}}_b + \bm{f}_k = \bm{Y}_{\Psi,b} \bm{\Psi}.
\end{equation}
We can also adjust the regressor and rewrite equation \eqref{eq: regressor nominal} to be a function of the reference velocity $\dot{\bm{q}}_r$ as follows:
\begin{equation} \label{eq: regressor1}
\tilde{\bm{H}}\ddot{\bm{q}}_r + \tilde{\bm{C}}\dot{\bm{q}}_r + \bm{f}_k = \bm{Y}_{\Psi,r} \bm{\Psi}
\end{equation}
In this form, the modified regressor $\bm{Y}_{\Psi,r}(\bm{q}_b, \dot{\bm{q}}_b, \dot{\bm{q}}_r, \ddot{\bm{q}}_r)$ depends on the reference velocity and acceleration, compared to the previous regressor $\bm{Y}_{\Psi,b}(\bm{q}_b, \dot{\bm{q}}_b, \ddot{\bm{q}}_b)$. Despite this change, both equations \eqref{eq: regressor nominal} and \eqref{eq: regressor1} employ the same vector of uncertainties, $\bm{\Psi}$. This new form, \eqref{eq: regressor1}, will aid in designing the adaptation law to ensure system stability.

Note that all uncertainties from the model, including friction, are combined into the vector \(\bm{\Psi}\), unlike in our previous work \cite{Sombolestan2023b}, where we had separate vectors for object dynamics model and friction properties. We found that having separate vectors of uncertainty did not significantly enhance the performance of the adaptive controller. Our primary focus is on compensating for the overall uncertainty of the system rather than distinguishing between different sources of uncertainty.

The unknown parameter updates according to the following adaptation laws:
\begin{equation}
\label{eq: adaptation law}
\dot{\hat{\bm{\Psi}}} = -\bm{\Gamma}_\Psi  {\bm{Y}_{\Psi,r}}^T \bm{s}
\end{equation}
where $\bm{\Gamma}_\Psi$ is a positive definite matrix. In \secref{sec: CLF}, we will show how the adaptation law \eqref{eq: adaptation law} ensures the system's stability when discussing our CLF constraint.

\subsection{Nonlinear MPC Problem} \label{sec: NMPC problem}
The motion planner problem is formulated as a nonlinear MPC problem consisting of a cost function and a set of nonlinear constraints. By defining these constraints, we ensure stability and safety for the team of robots during the manipulation task. These constraints are within the bounds provided for the inputs, such as geometric constraints of the object and manipulation force constraints. The general cost function for a nonlinear optimal control problem is as follows:
\begin{equation} \label{eq: cost}
J(\bm{u}[0, T],\bm{x}[0, T])= S(\bm{x}(T)) + \int_0^T l_s(\bm{x}(t), \bm{u}(t), t) \, dt
\end{equation}
where $S(.)$ is the final cost and $l_s(.)$ is the intermediate cost for a time horizon of $T$ and
subject to the dynamic and inequality constraints:
\begin{subequations} \label{eq: NLP st}
\begin{align}
    \dot{\bm{x}}(t) = \bar{f}(\bm{x},\hat{\bm{\Psi}}) + \bar{g}(\bm{x}) \bm{u} \label{eq: dynamic_eq}\\
    \bm{h}_{\text{bound}}(\bm{x},\bm{u}) \geq \bm{0} \label{eq: bound_eq} \\
    h_{\text{clf}}(\bm{x}, \bm{u}, \hat{\bm{\Psi}}) \geq 0 \label{eq: clf_eq} \\
    \bm{h}_{\text{cbf}}(\bm{x}, \bm{u}) \geq \bm{0} \label{eq: cbf_eq}
\end{align}
\end{subequations}
where the state vector includes:
\begin{equation} \label{eq: state vector}
    \bm{x} = [\bm{x}_b^T, \bm{d}^T]^T \in \mathbb{R}^{6+ N_r}
\end{equation}
and $\bm{d} = [d_{1}, \dots, d_{N_r}]^T$ represents contact point locations \( d_{i} \) for each of the \( N_r \) agents in the team.
The input vector is defined as:
\begin{equation} \label{eq: input vector}
\bm{u} = [\bm{F}_{r}^T, \dot{\bm{d}}^T]^T \in \mathbb{R}^{2 \times N_r}
\end{equation}
where $\bm{F}_{r} = [f_{r,1}, \dots, f_{r,N_r}]^T$ is a vector represents manipulation forces \( f_{r,i} \) and $\dot{\bm{d}} = [\dot{d}_{1}, \dots, \dot{d}_{N_r}]^T$ represent the rate of change of contact point locations. 
A detailed explanation of each component of our proposed optimal control problem, including \eqref{eq: cost} and \eqref{eq: NLP st}, will be provided in \secref{sec: NMPC framework}.

As described in \secref{sec: problem_statement} and illustrated in \figref{fig: object_schematic}, the manipulation force applied by each robot is perpendicular to the object's surface. Similarly, the contact point location is always tangential to the object's surface. Therefore, we only consider the force magnitude \( f_{r,i} \) and the distance \( d_i \) within the state and input vectors instead of treating them as a two-dimensional vector. The change in direction of these vectors is addressed within the dynamic equation, which will be further explained in \secref{sec: Dynamic Equation}. This approach reduces the dimension of the input vector for the motion planner MPC.
\section{Formulation of Nonlinear MPC Problem} \label{sec: NMPC framework}

This section provides a detailed explanation of the nonlinear MPC formulation. We will thoroughly discuss each component of the optimal problem introduced in \secref{sec: NMPC problem}, elaborating on the methods and considerations involved in their formulation.

\subsection{Cost Function} \label{sec: Cost Function}
The objective of the cost function is to minimize the deviation in object tracking from the desired path generated by the reference trajectory and to minimize the input vector. We propose the following quadratic cost function:

\begin{subequations} \label{eq: cost cont}
\begin{align}
l_s(\bm{x}(t), \bm{u}(t), t) = & \frac{1}{2}  \|\bm{x}_{b}(t) - \bm{x}_{b}^{\text{ref}}(t)\|_{\bm{Q}_{x_b}}  + \\ & \frac{1}{2} \|\bm{d}(t) - \bm{d}^*_{\text{prev}}\|_{\bm{Q}_d}  + \frac{1}{2} \|\bm{u}(t)\|_{\bm{R}_u} \nonumber \label{eq: cost l}\\
S(\bm{x}(T)) = & \frac{1}{2}  \|\bm{x}_{b}(T) - \bm{x}_{b}^{\text{ref}}(T)\|_{\bm{Q}_{f}}
\end{align} 
\end{subequations}
where $\|\mathbf{a}\|_{\bm{Q}}$ denotes the weighted norm $\mathbf{a}^T \bm{Q} \mathbf{a}$. The subgoal configurations $\bm{x}_{b}^{\text{ref}}$ are provided by the reference trajectory generator described in \secref{sec: Reference Trajectory}. All the matrices $\bm{Q}_f$, $\bm{Q}_{x_b}$, $\bm{Q}_d$, and $\bm{R}_u$ are positive definite.
For the contact point location $\bm{d}$, we want to minimize the change between the previous optimized solution and the current one, so the reference is $\bm{d}_{\text{prev}}^*$. This ensures a smooth change in the contact point location, leading to a more stable robot motion behavior. The last term in \eqref{eq: cost l} will aim to minimize the input value.
By including the terminal cost $S(\bm{x}(T))$, we place more weight on reaching the goal position rather than the intermediate goals.

\subsection{Dynamic Equation} \label{sec: Dynamic Equation}
The dynamic equation presented in the nonlinear MPC problem \eqref{eq: dynamic_eq}, in addition to considering the single rigid body dynamics, also addresses the constraint related to the interaction of each robot with the object. As mentioned previously, the robots are not connected rigidly to the object; therefore, they cannot directly apply torque to the object. Moreover, each robot can only apply force perpendicular to the object's surface. These constraints should be reflected in the system dynamic equation. Also, remember that there are model uncertainties in the dynamic model of the object, which we are going to compensate for using our proposed adaptive controller presented in \secref{sec: adaptive controller}. 

The dynamic equation with uncertainty \eqref{eq: uncertain_dynamic} can be presented in a control-affine form similar to \eqref{eq: dynamic_equation_control_affine} as follows:
 \begin{equation} \label{eq: dynamic_with_uncertainty}
 \dot{\bm{x}}_b = \bar{f}_b(\bm{x}_b)  +  \bar{g}_b(\bm{x}_b) \bm{\tau} + \left[ \begin{array}{c} \bm{0}_{3 \times 1} \\  -\bar{\bm{H}}(\bm{q}_b)^{-1}\bm{Y}_{\Psi,b} \bm{\Psi}  \end{array} \right]
\end{equation}
where:
\begin{subequations}
\begin{align}
 \bar{f}_b(\bm{x}_b) &= 
 \left[ \begin{array}{c} \dot{\bm{q}}_b \\  - \bar{\bm{H}}(\bm{q}_b)^{-1}\bar{\bm{C}}(\bm{q}_b, \dot{\bm{q}_b})\dot{\bm{q}_b} \end{array} \right]
  \\
 \bar{g}_b(\bm{x}_b) &= \left[ \begin{array}{c} \bm{0}_{3 \times 3} \\  \bar{\bm{H}}(\bm{q}_b)^{-1}  \end{array} \right]
\end{align}
\end{subequations}

The wrench vector $\bm{\tau} = [ \bm{F}^T, M_p]^T$ must be expressed as a function of the state vector \eqref{eq: state vector} and input vector \eqref{eq: input vector} of the nonlinear MPC problem. According to \figref{fig: object_schematic}, we have:
\begin{subequations}\label{eq: interaction constraint}
    \begin{align} 
   \bm{F} &= \bm{R}^T \sum_{i = 1 }^{N_r} \bm{f}_{r,i}  \\
 M_p &= \sum_{i = 1 }^{N_r} \bm{r}_i \times \bm{f}_{r,i}
    \end{align}
\end{subequations}
where
\begin{subequations}
\begin{align}
\bm{f}_{r,i} = f_{r,i} \hat{\bm{n}}_{r,i} \\
\bm{r}_i = \bm{r}_{i,o} + d_i {\hat{\bm{t}}_{r,i}}. 
\end{align}
\end{subequations}
The term \(\bm{r}_i \times \bm{f}_{r,i}\) represents the cross product of the two-dimensional vectors. The planar rotation matrix $\bm{R} \in \mathbb{R}^{2 \times2}$ as well as unit vectors $\hat{\bm{t}}_{r,i}$ and $\hat{\bm{n}}_{r,i}$ can be determined using the object state $\bm{x}_b$. The value of $f_{r,i}$ can be obtained from the input vector $\bm{u}$ as shown in \eqref{eq: input vector}. The vector $\bm{r}_{i,o}$ is a constant vector indicating the initial location of the contact point for each robot. 
Therefore, using the equations presented in \eqref{eq: interaction constraint}, the wrench vector $\bm{\tau}$ can be expressed as a function of the state vector and input vector, $\bm{\tau}(\bm{x}_b, \bm{u})$. 

Finally, by substituting the unknown uncertainty vector \(\bm{\Psi}\) with its estimated value \(\hat{\bm{\Psi}}\) in the dynamic equation \eqref{eq: dynamic_with_uncertainty}, rearranging the equations, and incorporating \(\dot{\bm{d}}\) as a part of the input vector, we can derive the adaptive dynamic equation as appeared in \eqref{eq: dynamic_eq}:
\begin{equation}
    \dot{\bm{x}}(t) = \bar{f}(\bm{x}, \hat{\bm{\Psi}}) + \bar{g}(\bm{x}) \bm{u}
\end{equation}

\subsection{State \& Input Boundaries} \label{sec: input boundaris}
The state and input boundaries presented in \eqref{eq: bound_eq} account for the allowable range for the optimized manipulation force and contact point locations. Since the robots are not rigidly attached to the object, the manipulation force applied by each robot always acts as a pushing force. Therefore, the magnitude of the optimized force must always be positive. Additionally, there is a maximum allowable value $F_{max}$ for the force that the robot can apply, which can depend on the robot's size. The boundaries for the manipulation force are as follows:
\begin{equation} \label{eq: force bound}
 0 \leq  f_{r,i} \leq F_{max}
\end{equation}

Furthermore, there are boundary limitations on $d_i$ to ensure that each robot remains in contact with the object's surface. The upper bound $\bar{d}_i$ and the lower bound $\underline{d}_i$ specify the limits of the object with respect to the origin $\bm{o}_i$ as shown in \figref{fig: object_schematic}. These boundaries can be formulated as follows:
\begin{equation} \label{eq: contact point bound}
\underline{d}_i \leq d_i \leq \bar{d}_i 
\end{equation}
We also impose a constraint on the rate of change \(\dot{d}_i\) in their contact point locations to ensure that the velocity of each agent  does not exceed a specified maximum value \(v_{max}\):
\begin{equation} \label{eq: vel_max}
    \|\dot{d}_i\| \leq v_{max}
\end{equation}

\subsection{CLF Constraint} \label{sec: CLF}

The CLF constraint in \eqref{eq: clf_eq} ensures the object's stability in tracking the desired trajectory by considering the adaptive dynamic \eqref{eq: dynamic_eq}. Let us examine the following Lyapunov candidate function:
\begin{equation} \label{eq: lyapunov function}
V(\bm{s}, \tilde{\bm{\Psi}}) = \frac{1}{2}(\bm{s}^T \bm{H} \bm{s} + \tilde{\bm{\Psi}}^{T} {\bm{\Gamma}_\Psi}^{-1} \tilde{\bm{\Psi}})
\end{equation}
where $\tilde{\bm{\Psi}} = \hat{\bm{\Psi}} - \bm{\Psi}$ represents the vector of estimation errors. Note that the inertia matrix $\bm{H}$ is positive definite. Since $\bm{\Psi}$ is a constant vector, the derivative of the estimation error $\dot{\tilde{\bm{\Psi}}}$ is the same as the derivative of the estimation $\dot{\hat{\bm{\Psi}}}$. Using this property, we can differentiate $V(t)$ as follows:
\begin{equation} \label{eq: lyapunov function differential}
\dot{V}(t) = \bm{s}^T \bm{H} \dot{\bm{s}} + \frac{1}{2}\bm{s}^T \dot{\bm{H}} \bm{s} + {\tilde{\bm{\Psi}}}^{T} {\bm{\Gamma}_\Psi}^{-1} \dot{\hat{\bm{\Psi}}}.
\end{equation}
Based on the definition of reference velocity in \eqref{eq: reference velocity}, we have $\dot{\bm{q}_b} = \bm{s} + \dot{\bm{q}}_r$ and $\dot{\bm{s}}= \ddot{\bm{q}_b} - \ddot{\bm{q}}_r$. Therefore, considering the equation of motion \eqref{eq: equation of motion}, the first two terms in equation \eqref{eq: lyapunov function differential} can be expanded as follows:
\begin{align}
\label{eq: expand}
&\bm{s}^T \bm{H} \dot{\bm{s}} + \frac{1}{2}\bm{s}^T \dot{\bm{H}} \bm{s} = \bm{s}^T \bm{H} (\ddot{\bm{q}_b} - \ddot{\bm{q}}_r) + \frac{1}{2}\bm{s}^T \dot{\bm{H}} \bm{s} = \nonumber \\
& \frac{1}{2}\bm{s}^T (\dot{\bm{H}} - 2\bm{C}) \bm{s} + \bm{s}^T [\bm{\tau} - (\bm{H}\ddot{\bm{q}}_r + \bm{C}\dot{\bm{q}}_r + \bm{f}_k)]
\end{align}
The term $\dot{\bm{H}} - 2\bm{C}$ is a skew-symmetric matrix \cite{Culbertson2021}, making $\bm{s}^T (\dot{\bm{H}} - 2\bm{C}) \bm{s}$ equal to zero. Furthermore, recall that the matrices $\bm{H}$ and $\bm{C}$ can be decomposed into nominal and unknown parts as described in \eqref{eq: uncertain_dynamic}, where $\bm{H} = \bar{\bm{H}} + \tilde{\bm{H}}$ and $\bm{C} = \bar{\bm{C}} + \tilde{\bm{C}}$. Using this decomposition and the definition in \eqref{eq: regressor1}, substituting \eqref{eq: expand} into \eqref{eq: lyapunov function differential} yields:
\begin{equation} \label{eq: lyapunov function differential 2}
\dot{V}(t) = \bm{s}^T [\bm{\tau} - \bar{\bm{H}}\ddot{\bm{q}}_r - \bar{\bm{C}}\dot{\bm{q}}_r - \bm{Y}_{\Psi,r} \bm{\Psi}] + {\tilde{\bm{\Psi}}}^{T} {\bm{\Gamma}_\Psi}^{-1} \dot{\hat{\bm{\Psi}}}.
\end{equation}
Finally, substituting the adaptation law \eqref{eq: adaptation law} into equation \eqref{eq: lyapunov function differential 2} gives:
\begin{equation} \label{eq: lyapunov function differential 3}
\dot{V}(t) = \bm{s}^T [\bm{\tau} - \bar{\bm{H}}\ddot{\bm{q}}_r - \bar{\bm{C}}\dot{\bm{q}}_r - \bm{Y}_{\Psi,r} \hat{\bm{\Psi}}].
\end{equation}
According to the Lyapunov theorem \cite{Slotine1991}, if we can ensure that $\dot{V}(t) \leq 0$, the system will be uniformly stable because $V(t)$ is positive definite and decrescent, and $\dot{V}(t)$ is negative semi-definite. As a result, the variables $\bm{s}$ and $\tilde{\bm{\Psi}}$ will remain bounded.

Additionally, let's define a positive definite function $W(\bm{s})$ as follows:
\begin{equation}
W(\bm{s}) := \frac{1}{2} \bm{s}^T \bm{K}_D \bm{s}
\end{equation}
and assume $\dot{V}(t) + W(\bm{s}) \leq 0$. Since $\ddot{V}(t) + \dot{W}(t)$ is bounded and $\dot{V}(t) + W(\bm{s})$ is uniformly continuous in time, and $V(t)$ is lower bounded, the second version of Barbalat's Lemma \cite{Slotine1991} implies that $\dot{V}(t) + W(\bm{s}) \rightarrow 0$ as $t \rightarrow \infty$, which means $W(\bm{s}) \rightarrow 0$. Therefore, $\bm{s}$ also approaches zero as $t \rightarrow \infty$. When $\bm{s} = 0$, it can be shown that $\dot{\bm{q}}_e = - \lambda \bm{q}_e$ according to the definition of the composite error \eqref{eq: composite error}, which corresponds to an asymptotically stable system.

To achieve the asymptotic stability described above, we can include the required property $\dot{V}(t) + W(\bm{s}) \leq 0$ as a constraint within our motion planner. Thus, we formulate the CLF constraint as follows:
\begin{align}
h_{\text{clf}}(\bm{x}, \bm{u}, \hat{\bm{\Psi}}) := &\bm{s}^T [-\bm{\tau} + \bar{\bm{H}}\ddot{\bm{q}}_r + \bar{\bm{C}}\dot{\bm{q}}_r + \bm{Y}_{\Psi,r} \hat{\bm{\Psi}}] - \nonumber \\
& \frac{1}{2} \bm{s}^T \bm{K}_D \bm{s}
\end{align}
By choosing a proper input value $\bm{u}$, which generates the value $\bm{\tau}(\bm{x}_b, \bm{u})$ as explained in \eqref{eq: interaction constraint}, we can achieve the CLF constraint $h_{\text{clf}} \geq 0$ as presented in \eqref{eq: clf_eq}.

\subsection{CBF Constraints} \label{sec: CBF Constraints}
In most practical scenarios, the environment is filled with obstacles, which can sometimes move, such as when humans are present. Ensuring safety is crucial for any loco-manipulation task, involving both collision avoidance between objects and obstacles and preventing collisions between robots and obstacles. We leverage the flexibility of each robot's movement to adjust the contact point location, enabling the object to follow the desired path while coordinating each robot to avoid obstacles.

Considering \(N_o\) obstacles, as mentioned in the problem statement, for each obstacle \(o_j\), there is an associated barrier function for the manipulated object (\(B_{o_j}^{m} > 0\)), which defines the safe boundary around the obstacles. This concept is also applied to the potential collisions between each obstacle and each robot \(r_i\) from all \(N_r\) agents, defining the barrier functions as \(B_{o_j}^{r_i} > 0\). Each barrier function can be defined as the boundary of the safe set for spherical obstacles as follows:
\begin{subequations} \label{eq: barrier functions}
\begin{align}
 B_{o_j}^m(\bm{x}) = \|\bm{\mathcal{O}}_j - \bm{x}_p \| - R_{j,m} \label{eq: barrier functions object}
 \\
 B_{o_j}^{r_i}(\bm{x}) = \|\bm{\mathcal{O}}_j - \bm{\mathcal{R}}_i \| - R_{j,r_i} \label{eq: barrier functions robots}
\end{align}
\end{subequations}
where \(\bm{\mathcal{O}}_j\) and \(\bm{\mathcal{R}}_i\) represent the positions of each obstacle and the robot's center, respectively. \(R_{j,m}\) is the obstacle's barrier radius accounting for the size of the manipulated object, and \(R_{j,r_i}\) is the obstacle's barrier radius accounting for the size of robot \(r_i\).

The barrier functions defined in \eqref{eq: barrier functions object} can be used as CBFs of relative degree 2 since it depend only on \(\bm{x}_p\). The barrier functions defined in \eqref{eq: barrier functions robots} can be used as CBFs of relative degree 1 since \(\bm{\mathcal{R}}_i\) depends on \(d_i\). According to definition \ref{def:ecbf}, which defines ECBFs, we will have:
\begin{subequations} \label{eq: cbfs}
\begin{align}
\ddot{B}_{o_j}^m(\bm{x}, \bm{u}) + \beta_{o_j}^{m} \dot{B}_{o_j}^{m}(\bm{x}) + \alpha_{o_j}^{m} B_{o_j}^{m}(\bm{x}) \geq 0 \label{eq: cbf1} \\
\dot{B}_{o_j}^{r_i}(\bm{x}, \bm{u}) + \alpha_{o_j}^{r_i} B_{o_j}^{r_i}(\bm{x}) \geq 0 \label{eq: cbf2}
\end{align}
\end{subequations}
Note that when computing the Lie derivative to obtain \(\ddot{B}\) and \(\dot{B}\), the adaptive dynamic equation \eqref{eq: dynamic_eq} will be used. The parameters \(\beta_{o_j}^{m}\), \(\alpha_{o_j}^{m}\), and \(\alpha_{o_j}^{r_i}\) should be chosen such that the roots of equations \eqref{eq: cbf1} and \eqref{eq: cbf2} are negative real values \cite{Nguyen2016ExponentialConstraints}. Thus, the equations \eqref{eq: cbfs} will form the CBF constraints \eqref{eq: cbf_eq} within the motion planner.


\subsection{Penalty Cost} \label{sec: penalty cost}
All the inequality constraints in the nonlinear optimal control problem \eqref{eq: NLP st}, including \( h_{bound} \), \( h_{clf} \), and \( h_{cbf} \), will be formulated as penalty costs and incorporated into the cost function. This approach is commonly used in nonlinear MPC problems with numerous constraints to simplify the numerical solver's task \cite{Grandia2022a}. In this work, we will utilize relaxed barrier functions \cite{Hauser2006AConstraints, Feller2017AFunctions}. The penalty cost for each inequality constraint \( h \) is determined using two positive scalar variables, \(\rho\) and \(\epsilon\), as follows:
\begin{equation} \label{eq: penalty cost function}
 \mathcal{P}(h) = 
\begin{cases}
    - \rho \ln(h), & \text{if } h \geq \epsilon, \\
    \frac{\rho}{2} \left( \left( \frac{h - 2\epsilon}{\epsilon} \right)^2 - 1 \right) - \rho \ln(\epsilon), & \text{if } h < \epsilon.
\end{cases}   
\end{equation}
This function acts as a log-barrier in the feasible region (\( h \geq \epsilon \)), and transitions to a quadratic function when the constraint value is within a distance \( h < \epsilon \). The parameter \( \rho \) scales the penalty cost, and the values of \( \epsilon \) and \( \rho \) can be chosen based on the sensitivity of each constraint, ensuring stability and safety while optimizing solver performance. 

For all inequality constraints, the combined penalty cost is defined as:
\begin{equation}
l_{\mathcal{P}}(\bm{x}, \bm{u}, t) = \sum_{i \in \mathcal{I}} \mathcal{P}(h_\text{bound}^i) + \mathcal{P}(h_\text{clf}) + \sum_{j \in \mathcal{J}} \mathcal{P}(h_\text{cbf}^j)
\end{equation}
where \(\mathcal{I}\) and \(\mathcal{J}\) represent the sets of all state/input boundaries \eqref{eq: bound_eq} and CBF constraints \eqref{eq: cbf_eq}, respectively. The combined penalty cost \( l_{\mathcal{P}} \) will be added to the intermediate cost \( l_{s} \) in the cost function \eqref{eq: cost}, substituting the inequality constraints.

\subsection{Solving the Nonlinear MPC Problem}
The continuous control input \eqref{eq: input vector} is parameterized over subintervals of the prediction horizon $[t, t + T]$ to convert it into a finite-dimensional decision problem. This is discretized into $N_T$ steps with $k \in \{0, \dots, N_T-1\}$, where the time step is \(\Delta t = T / N_T\) and the control time is defined as \(t_k = t + \Delta t \times k\). Using a discretization method like zero-order hold \cite{Fadali2012}, we parameterize the state and input as \(\bm{x}_{k+1} = \bm{x}(t_{k+1})\) and \(\bm{u}_{k} = \bm{u}(t_k)\). Thus, the discretized nonlinear MPC problem is formulated as follows:
\begin{subequations} \label{eq: discrete NMPC}
\begin{align}
\min_{\bm{X,U}} \quad & \|\bm{x}_{b,N_T} - \bm{x}_{b,N_T}^{\text{ref}}\|_{\bm{Q}_f} +  \sum_{k=0}^{N_T-1} \|\bm{x}_{b,k+1} - \bm{x}_{b,k+1}^{\text{ref}}\|_{\bm{Q}_{x_b}} \nonumber \\
& + \|\bm{d}_{k+1} - \bm{d}^*_{\text{prev}}\|_{\bm{Q}_d} + \|\bm{u}_k\|_{\bm{R}_u}, \\
\text{s.t.} \quad & \bm{x}_{k+1} = \bar{f}^k(\bm{x}_k, \hat{\bm{\Psi}}) + \bar{g}^{k}(\bm{x}_k) \bm{u}_k, \\
& \bm{h}_{\text{bound}}^k(\bm{x}_k,\bm{u}_k) \geq 0, \\
& h_{\text{clf}}^k(\bm{x}_k, \bm{u}_k, \hat{\bm{\Psi}}) \geq 0, \\
& \bm{h}_{\text{cbf}}^k(\bm{x}_k, \bm{u}_k) \geq 0,
\end{align}
\end{subequations}
where \(\bm{X} = [\bm{x}_0^T, \dots , \bm{x}_{N_T}^T]^T\) and \(\bm{U} = [\bm{u}_0^T, \dots , \bm{u}_{N_T}^T]^T\) are the vectors of discretized state and input over the prediction horizon, and \(\bar{f}^k\), \(\bar{g}^k\), and \(\bar{h}_{(.)}^k\) represent the discrete samples of their continuous counterparts. Considering the penalty cost instead of the inequality constraints described in \secref{sec: penalty cost}, the discretized nonlinear MPC problem \eqref{eq: discrete NMPC} can be expressed as follows:
\begin{subequations}
\begin{align}
\min_{\bm{X,U}} \quad & S(\bm{x}_{N_T}) + \sum_{k=0}^{N_T-1} l_s^k(\bm{x}_{k}, \bm{u}_{k}) + l_{\mathcal{P}}^k(\bm{x}_{k}, \bm{u}_{k}) \\
\text{s.t.} \quad & \bm{x}_{k+1} = \bar{f}^k(\bm{x}_k, \hat{\bm{\Psi}}) + \bar{g}^{k}(\bm{x}_k) \bm{u}_k,
\end{align}
\end{subequations}
where \(l_s^k\) and \(l_\mathcal{P}^k\) are the discretized versions of their continuous counterparts. The nonlinear MPC problem can then be reformulated as a general nonlinear problem (NLP) by augmenting the decision variables as \(\bm{\mathcal{X}} = [\bm{X}^T, \bm{U}^T]^T\):
\begin{subequations} \label{eq: NLP}
\begin{align}
\underset{\bm{\mathcal{X}}}{\min} & \quad \bm{\Phi}(\bm{\mathcal{X}}), \\ 
\text{s.t.} & \quad \bm{\mathcal{F}}(\bm{\mathcal{X}}) = \bm{0},
\end{align}
\end{subequations}
where \(\bm{\Phi}(\bm{\mathcal{X}})\) is the cost function and \(\bm{\mathcal{F}}(\bm{\mathcal{X}})\) represents the dynamic equation constraints.

To efficiently solve the proposed NLP \eqref{eq: NLP}, we linearize the problem and solve it iteratively using a sequential quadratic programming (SQP) approach \cite{Nocedal2006NumericalOptimization}. Given the current iteration solution \(\bm{\mathcal{X}}_i\), we compute the deviation \(\delta \bm{\mathcal{X}}\). The problem \eqref{eq: NLP} can then be approximated by a quadratic programming (QP) problem with respect to \(\delta \bm{\mathcal{X}}\) as follows:
\begin{subequations}
\begin{align}
\underset{\delta \bm{\mathcal{X}}}{\min} & \quad {\nabla_{\bm{\mathcal{X}}} \bm{\Phi}(\bm{\mathcal{X}}_i)}^T \delta \bm{\mathcal{X}} + \frac{1}{2} {\delta \bm{\mathcal{X}}}^T  \bm{\mathcal{H}}_i  \delta \bm{\mathcal{X}}, \\ 
\text{s.t.} & \quad \bm{\mathcal{F}}(\bm{\mathcal{X}}_i) +  {\nabla_{\bm{\mathcal{X}}} \bm{\mathcal{F}}(\bm{\mathcal{X}}_i)}^T \delta \bm{\mathcal{X}} = \bm{0},
\end{align}
\end{subequations}
where \(\bm{\mathcal{H}}_i = \nabla_{\bm{\mathcal{X}}}^2 \bm{\Phi}(\bm{\mathcal{X}}_i)\) is the Hessian matrix. If the Hessian is positive semi-definite, the QP is convex and can be solved efficiently. All the costs defined in \eqref{eq: cost cont}, including the intermediate cost \(l_s(.)\) and the terminal cost \(S(.)\), are already in quadratic form, satisfying the Hessian requirements. For any penalty cost, we can approximate the following to exploit the convexity for the Hessian:
\begin{equation}
\nabla_{\bm{\mathcal{X}}}^2 \left( \mathcal{P}(\bm{h}(\bm{\mathcal{X}})) \right) \approx \nabla_{\bm{\mathcal{X}}} \bm{h}(\bm{\mathcal{X}})^T \nabla_{\bm{h}}^2 \mathcal{P}(\bm{h}(\bm{\mathcal{X}})) \nabla_{\bm{\mathcal{X}}} \bm{h}(\bm{\mathcal{X}}),
\end{equation}
where \(\nabla_{\bm{h}}^2 \mathcal{P}(\bm{h}(\bm{\mathcal{X}}))\) is a diagonal and positive definite matrix according to the definition of penalty cost \(\mathcal{P}(h)\) in \eqref{eq: penalty cost function}.
\section{Results} \label{sec: results}
\begin{figure}[t!]
	\centering
	\subfloat[Early state of the manipulation task]{\includegraphics[width=1\linewidth]{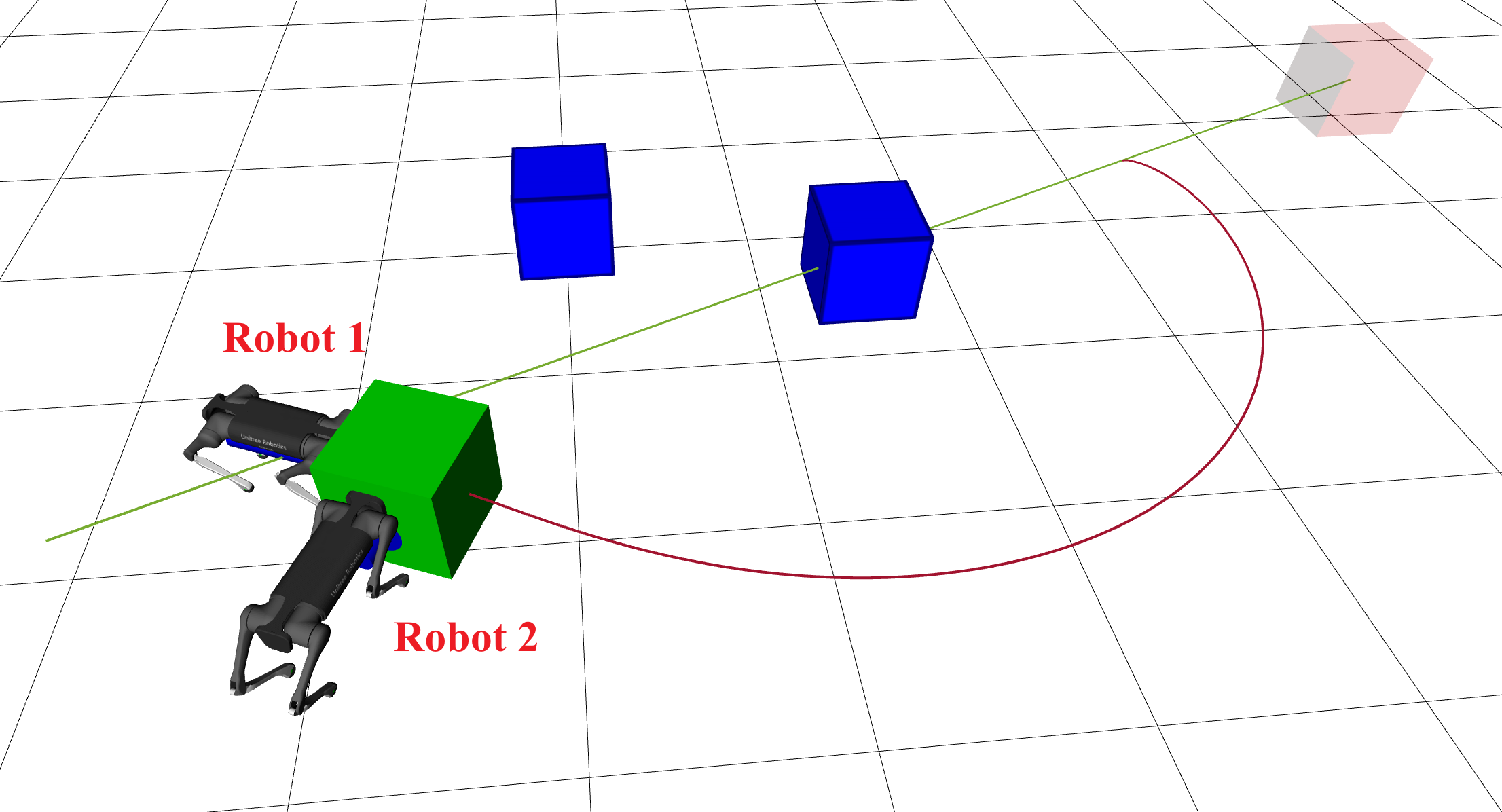}}
	\hfill
	\subfloat[Motion planner with adaptive controller successfully completes the task]{\includegraphics[width=0.48\linewidth]{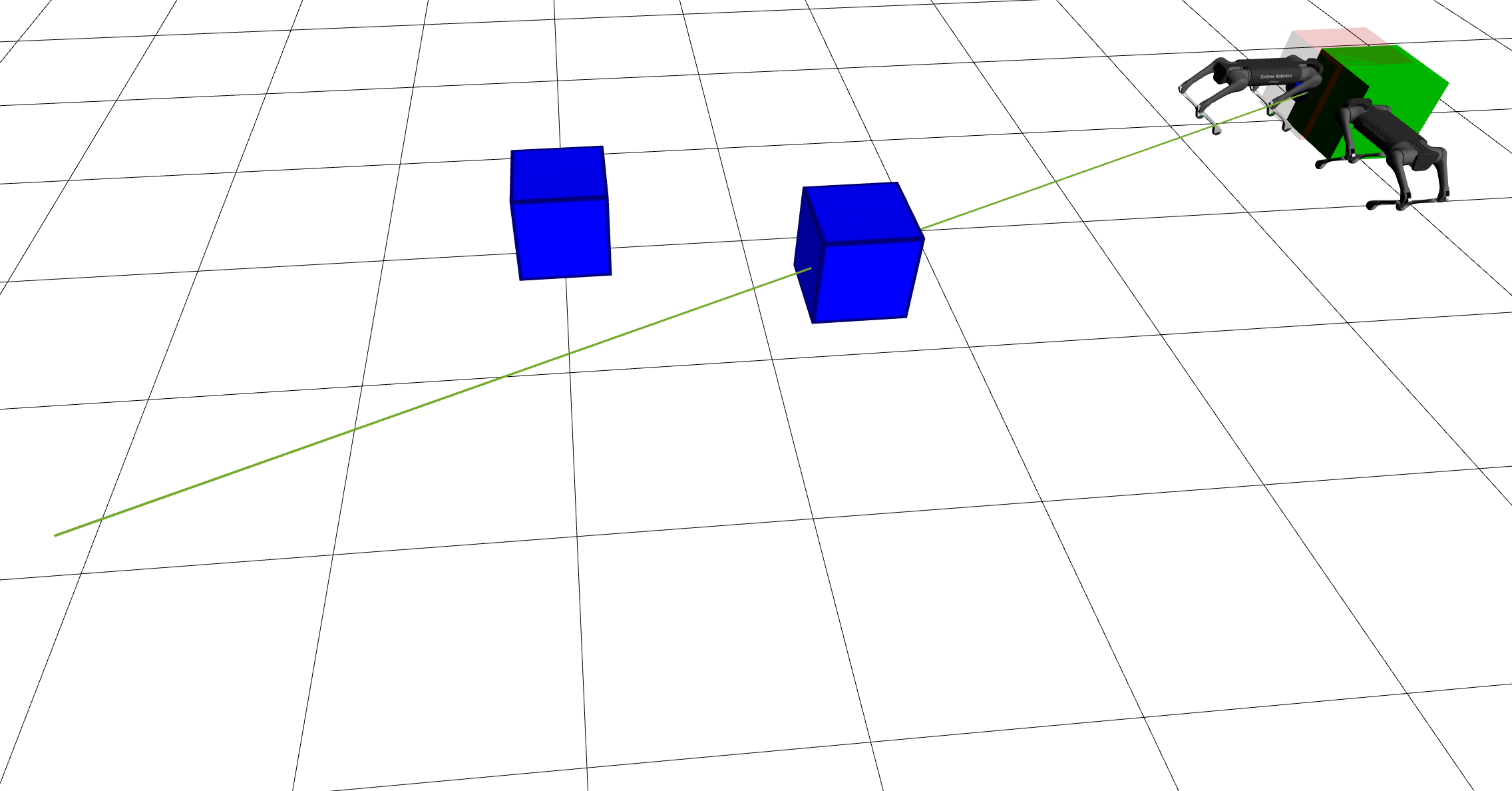}}
    \hfill
    \subfloat[Motion planner without adaptive controller gets stuck halfway]{\includegraphics[width=0.48\linewidth]{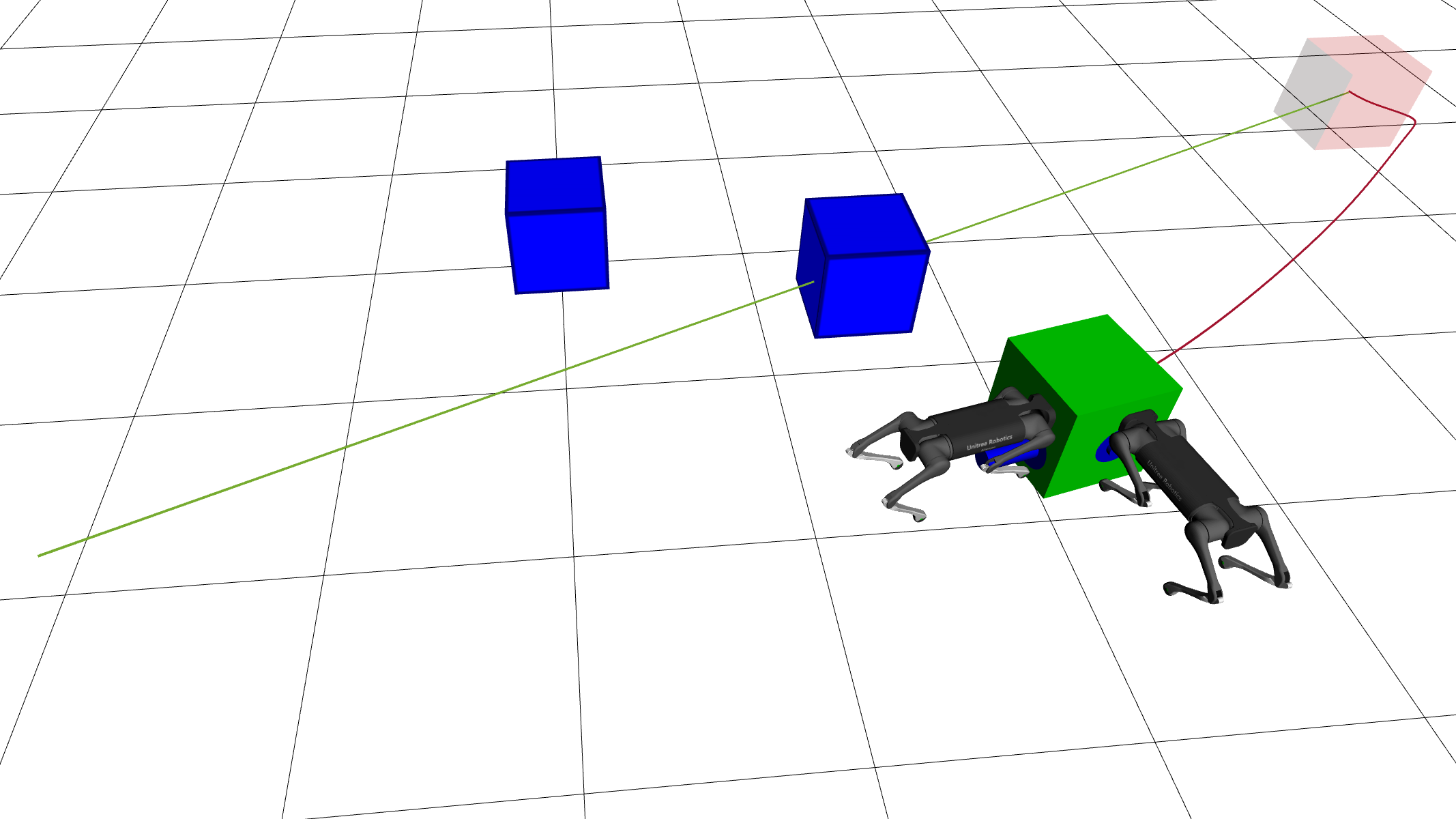}}
	\hfill
	\subfloat[Manipulation force plot for robot 1 ($f_{r,1}$)]{\includegraphics[width=1\linewidth]{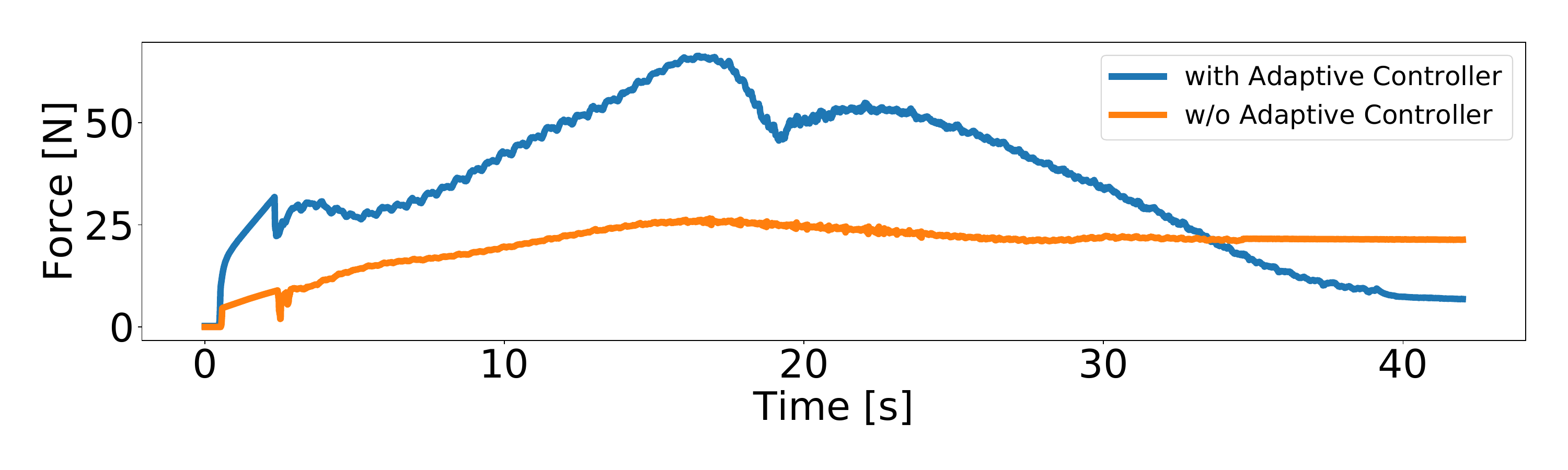}}
    \hfill
	\subfloat[Manipulation force plot for robot 2 ($f_{r,2}$)]{\includegraphics[width=1\linewidth]{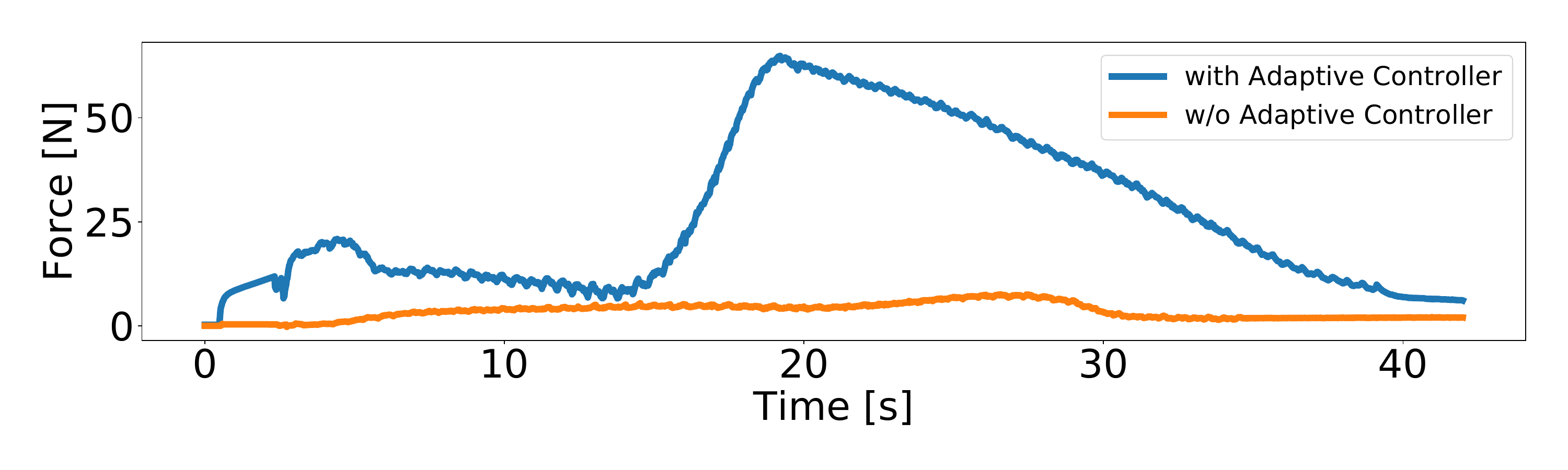}}	
	\caption{\textbf{Comparing the performance of the motion planner with and without the adaptive controller.} In the snapshots, the green box is the manipulated object, the red cube is the user-defined target location, and the two blue boxes are static obstacles. The green line represents the straight path from the initial position to the target position of the manipulated object, while the red line indicates the optimized path from the motion planner, considering safety and other constraints.}
	\label{fig: comparison adaptive}
\end{figure}

To evaluate the performance of the proposed controller, we present results across various scenarios, both in simulation and on hardware. For the simulation, we use the Gazebo simulator along with ROS 1 Noetic. All parameters in the planner and decentralized loco-manipulation controllers remain constant throughout the simulations and hardware experiments. To implement the motion planner's nonlinear MPC problem, we use the OCS2 package \cite{Farshidian2017OCS2:Systems}. A time horizon of \( T = 5 \) seconds is used for the planner's MPC problem, which is updated at 100 Hz. The rest of the parameters are listed in \tabref{tab: parameter}. These parameters are initially selected in simulation and fine-tuned during hardware experiments. Further details of the results can be found in the supplementary video accompanying this paper \footnote{\href{https://youtu.be/cU_qevkW86I}{https://youtu.be/cU\_qevkW86I}}.

\begin{table}[bt!]
	\caption{Motion Planner Settings}
	\label{tab: parameter}
	\begin{tabular}{l l}
		\bf{Parameter} & \bf{Value} \\ 
        \hline \hline
        \multicolumn{2}{l}{Reference Trajectory} \\ \hline
        $v_{\text{avg}}$ & 0.5 [m/s] \\
        $\omega_{\text{avg}}$ & 0.8 [rad/sec] \\ 
        \hline 
        \multicolumn{2}{l}{Adaptive Controller} \\ \hline
        $\bm{\Gamma}_{\Psi}$ & diag$(3, 2, 1, 1) \times 10^2$ \\
        $\lambda$ & 3 \\
        \hline
        \multicolumn{2}{l}{MPC Cost Function} \\ \hline
        $\bm{Q}_f$ & diag$(150, 150, 3, 3, 3, 8)$ \\
        $\bm{Q}_{\bm{x}_b}$ & diag$(20, 22, 2, 3, 3, 1)\times 10^{-1}$ \\
        $\bm{Q}_{d}$ & $\bm{I}_{N_r} \times 10^{-1}$ \\
        $\bm{R}_u$ & $\bm{I}_{2 \times N_r} \times 10^{-2}$ \\
        \hline
        \multicolumn{2}{l}{MPC Constraints} \\ \hline
        ($F_{max}, v_{max}$) & (0.7 [N], 1 [m/s])\\ 
        $(\alpha, \beta)_{\text{CBF}}$ & (4, 4) \\ 
        $(\rho, \epsilon)_{\text{CBF}}$ & (0.8, 0.5) \\ 
        $(\rho, \epsilon)_{\text{CLF}}$ & (1, 0.5) \\ 
        $(\rho, \epsilon)_{\text{bound}}$ & (0.1, 0.01) \\
        $\bm{K}_{D}$ & $3\bm{I}_{3}$
	\end{tabular}
\end{table}

For the decentralized loco-manipulation control setup, we use the exact control system implementation as presented in \cite{Sombolestan2023b}. Given that the loco-manipulation MPC was linearly formulated through specific assumptions \cite{Sombolestan2023b}, we leverage this linearity to formalize the MPC problem as a QP problem and use the qpOASES package \cite{Ferreau2014QpOASES:Programming} as the solver. Our code is accessible in an open-source repository \footnote{\href{https://github.com/DRCL-USC/collaborative_loco_manipulation}{https://github.com/DRCL-USC/collaborative\_loco\_manipulation}}.

\begin{figure}[t!]
\centering
\subfloat[First initial configuration]{\includegraphics[width=0.46\linewidth]{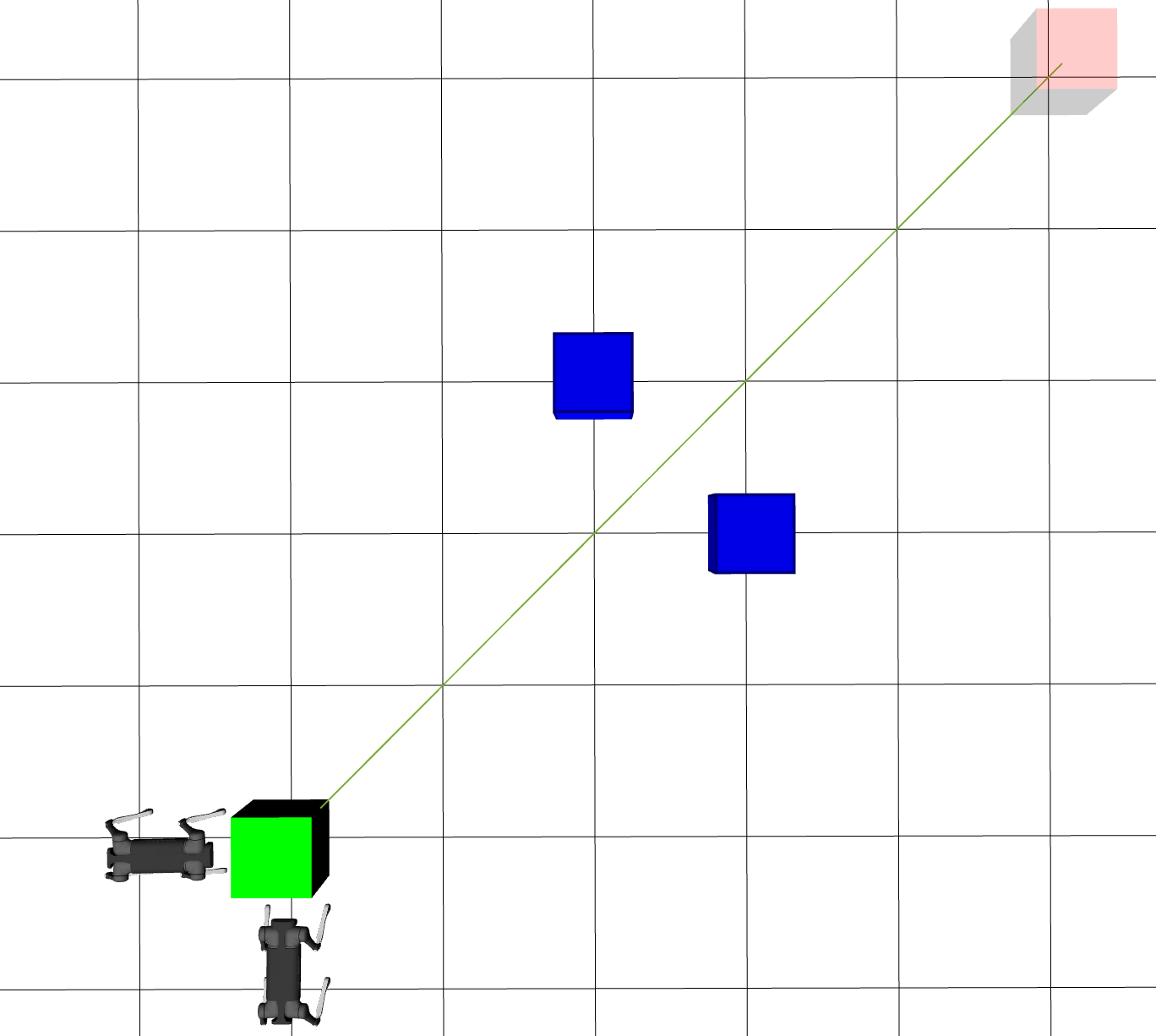} \label{fig: from left}}
\hfill
\subfloat[Optimized trajectory with the first configuration]{\includegraphics[width=0.48\linewidth]{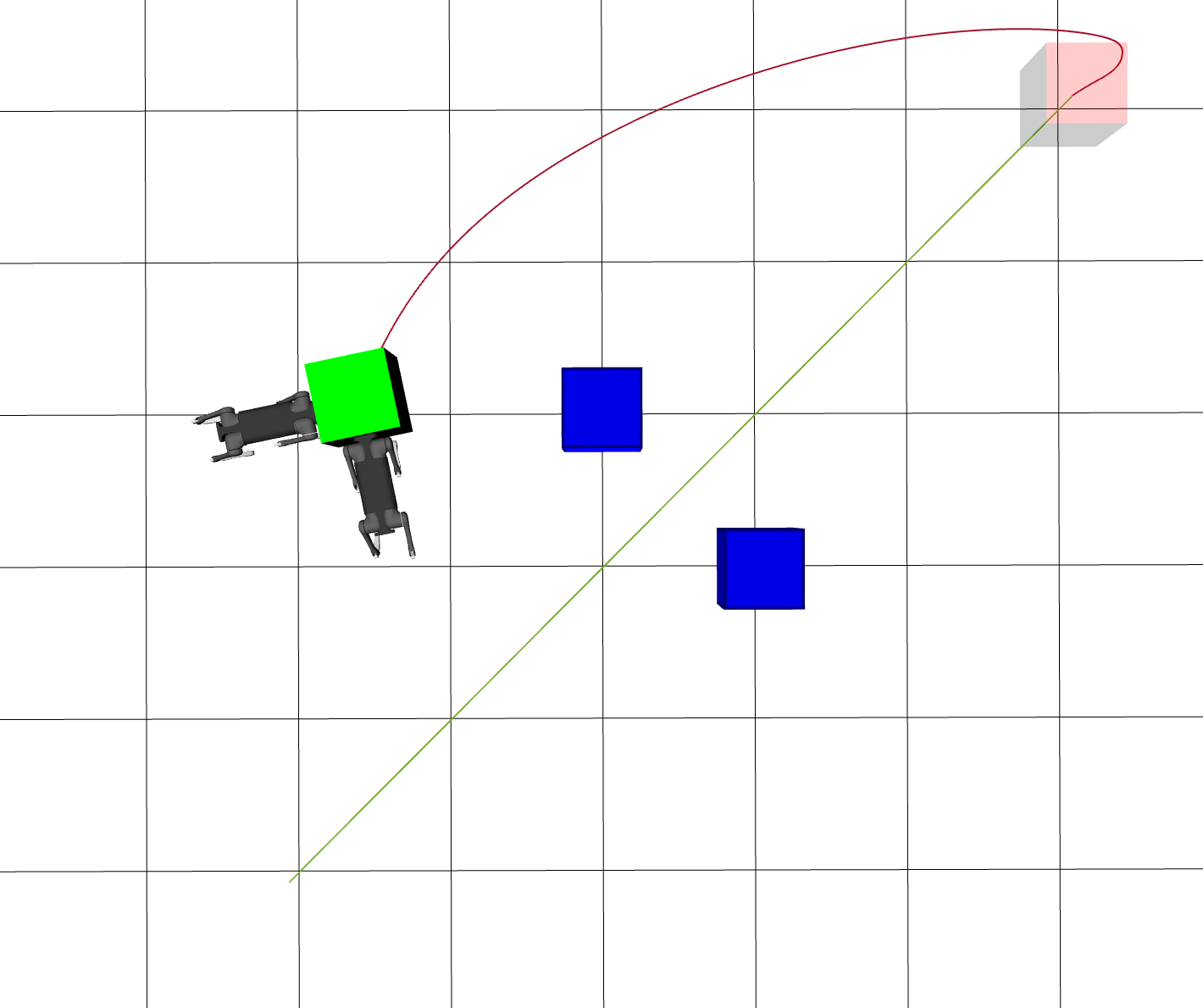} \label{fig: from left middle}}
\hfill
\subfloat[Second initial configuration]{\includegraphics[width=0.48\linewidth]{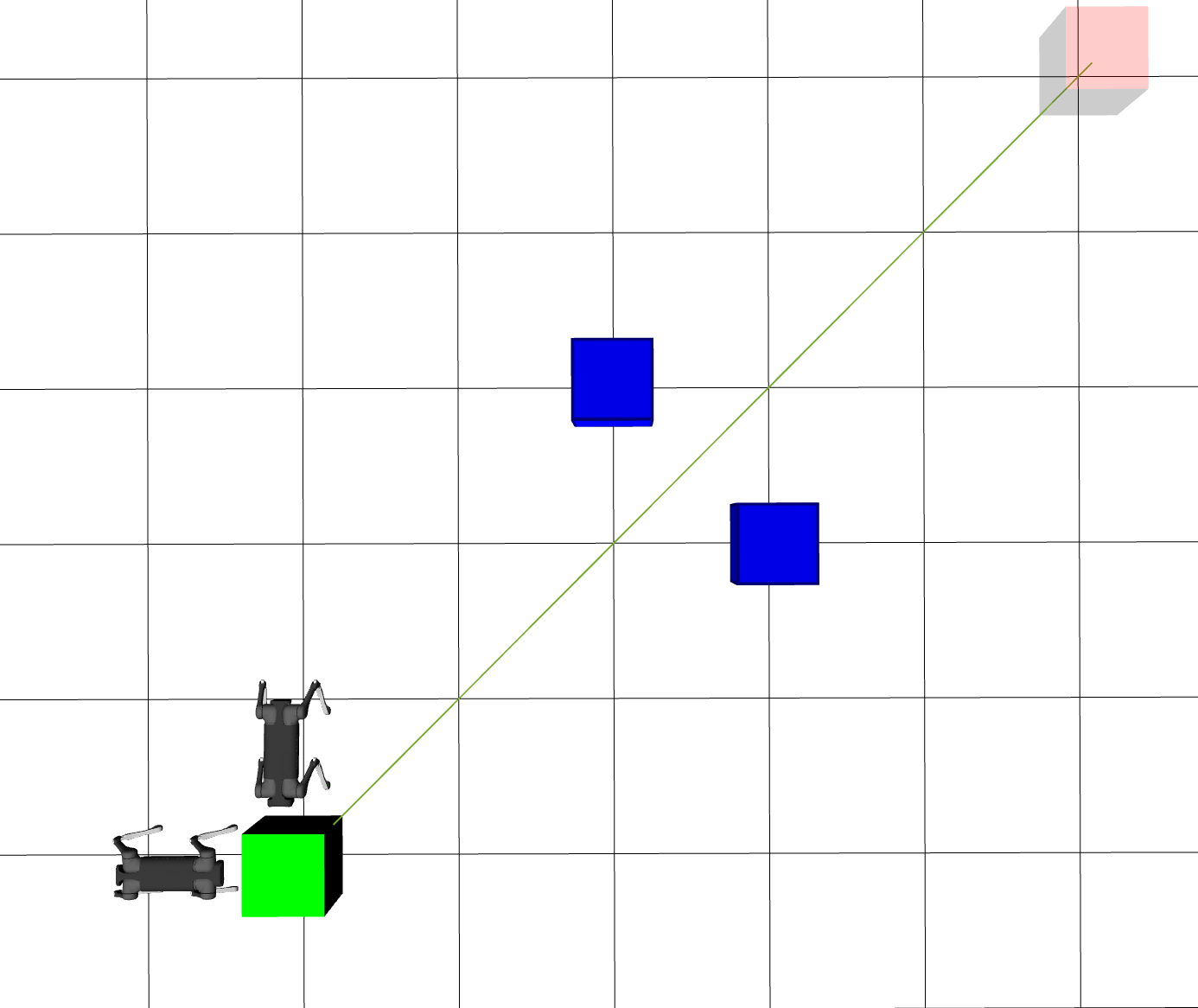} \label{fig: from right}}
\hfill
\subfloat[Optimized trajectory with the second configuration]{\includegraphics[width=0.43\linewidth]{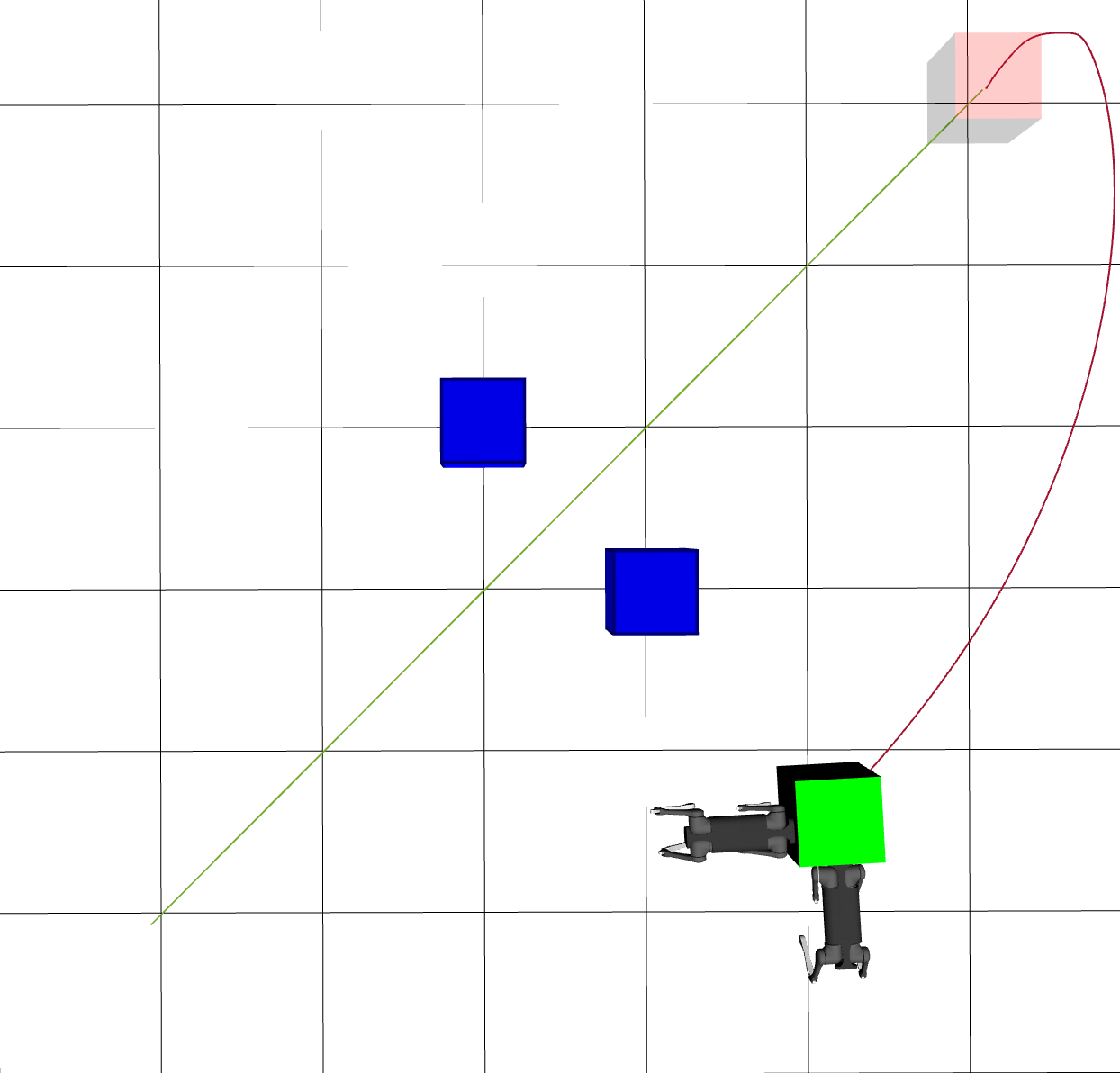} \label{fig: from right middle}}
\caption{\textbf{Impact of Initial Robot Configuration on Optimized Trajectories.} (a) and (b) show the first initial configuration and its resulting optimized trajectory, where the robots navigate the object from the left side of the obstacles. (c) and (d) display the second initial configuration and its resulting optimized trajectory, where the robots maneuver the object through the right side of the obstacles.}
\label{fig: comparison initial config}
\end{figure}

\subsection{Comparative Analysis}
To demonstrate the effectiveness of our motion planner structure, we perform a comparative analysis to examine our proposed method. All these simulation designs are kept as simple as possible, focusing only on the effects of specific components in our motion planner.
\subsubsection{Effect of Adaptive Controller}:
For this part, we conduct two identical simulations with two Aliengo robots attempting to manipulate a cubic object in an environment with two static obstacles, aiming to reach a target point. In one simulation, we use our proposed method; in the other, we disable the adaptive controller. In both simulations, the controller assumes the object's mass to be 6 kg, while the actual mass is 8 kg. Additionally, we do not include friction in the dynamic equation, meaning the controller assumes a frictionless contact between the object and the ground. However, in the simulation, we introduce friction with a coefficient of 0.4 between the object and the ground.

The robots successfully manipulate the object to the target location with the adaptive controller. Without the adaptive controller, the robots get stuck midway, unable to push the object further because the force provided by the planner is insufficient to overcome the friction. The force plots of each robot in both scenarios are illustrated in \figref{fig: comparison adaptive}. As shown, the adaptive dynamic helps the planner account for uncertainties and compensate for them in each robot's force, which is significant given the large uncertainties in the system.

\subsubsection{Effect of CBFs for Collision Avoidance of Each Robot}:
In this section, we examine the impact of CBFs on each robot's collision avoidance with obstacles. We conduct two identical simulations: one using our proposed method and the other without the CBFs related to robots and obstacles. The simulation setup is the same as in the previous part, where we examined the adaptive controller.

\begin{figure}[t!]
\centering
\subfloat[Motion planner utilizing CBF constraints for individual robot collision avoidance ($B_{o_j}^{r_i}$)]{\includegraphics[width=0.8\linewidth]{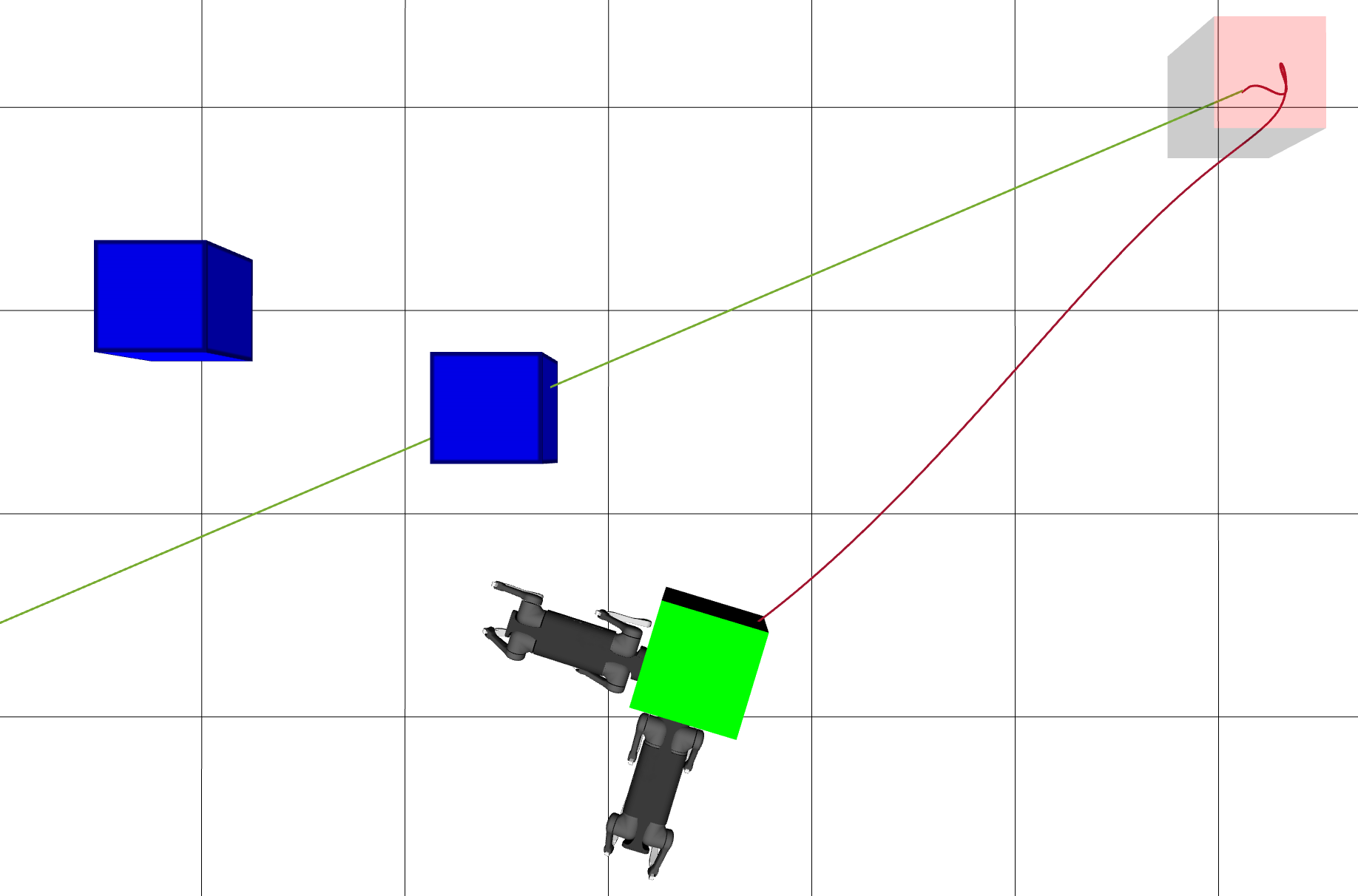}}
\hfill
\subfloat[Motion planner without CBF constraints for individual robot collision avoidance]{\includegraphics[width=0.8\linewidth]{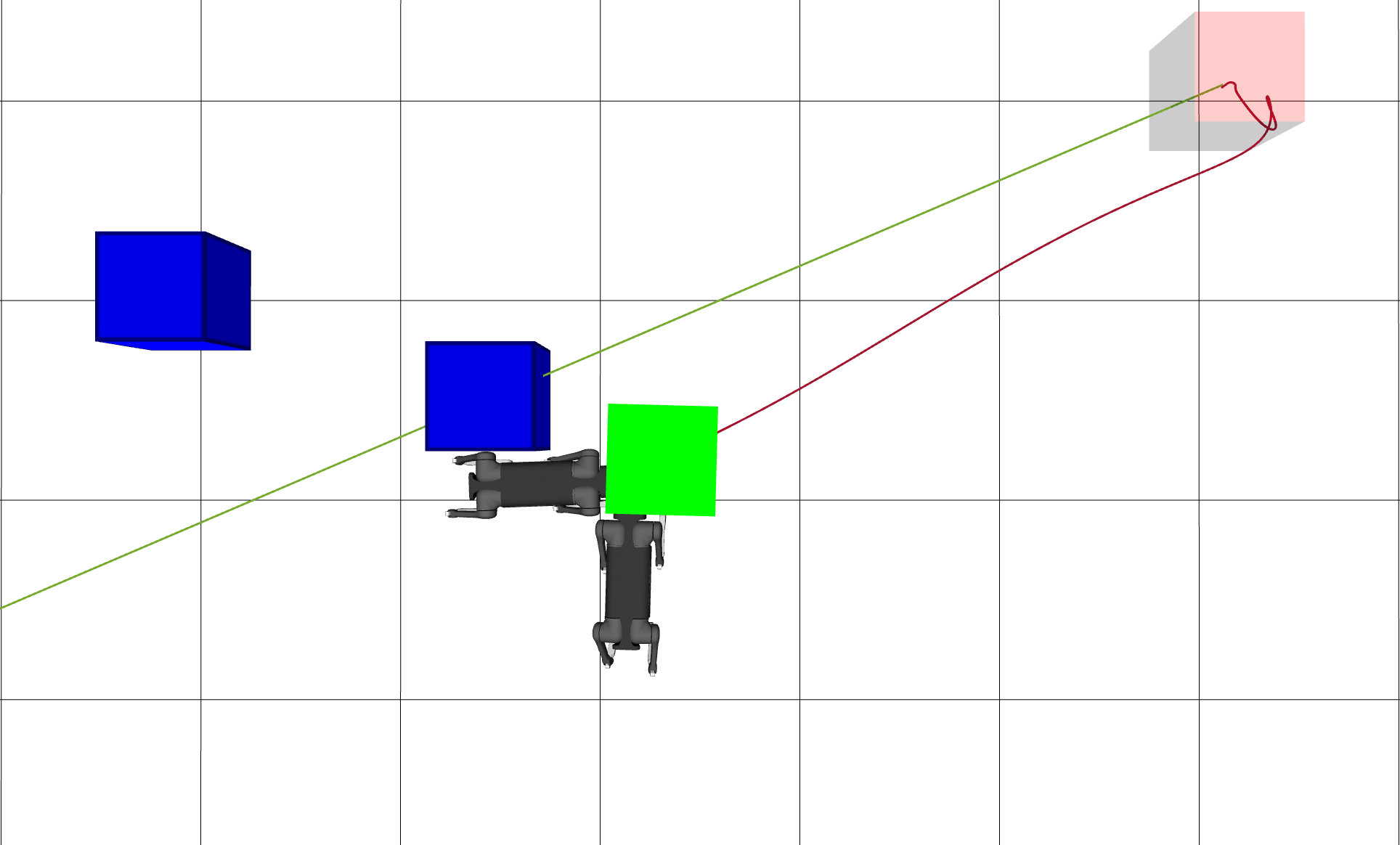}}
\caption{\textbf{Performance comparison of the motion planner with and without CBF constraints for robot collision avoidance.} In snapshot (a), the robot maintains a safe distance from obstacles due to the CBF constraints in the motion planner. In snapshot (b), the robot collides with the obstacle due to the absence of CBF constraints.}
\label{fig: comparison cbf}
\end{figure}

Note that in the motion planner, we still include CBFs for the collision avoidance of the manipulated object and obstacles, removing only those for each robot. Therefore, the planner still optimizes the trajectory to prevent collisions between the manipulated object and obstacles. However, as shown in \figref{fig: comparison cbf}, the robots collide with obstacles without CBFs associated with the robots and obstacles. In contrast, using our proposed method, the entire system navigates the environment safely, avoiding any collisions with obstacles.

\begin{figure*}[t!]
\centering
\subfloat[Start point]
{\includegraphics[width=0.3\linewidth]{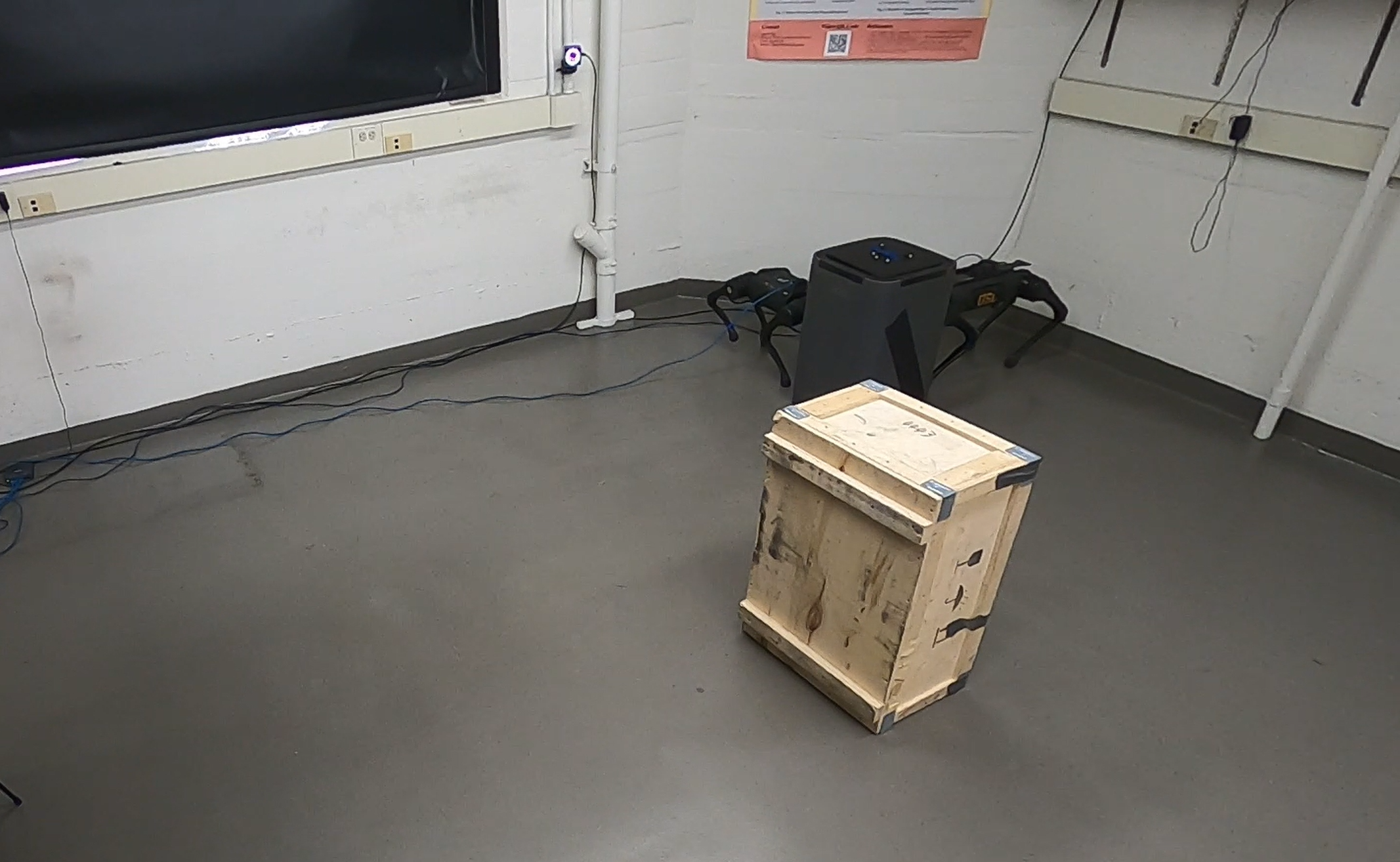}}
\hfill
\subfloat[Maintaining safe distance from obstacle]
{\includegraphics[width=0.3\linewidth]{Figures/Experiment/5.png}}
\hfill
\subfloat[Reaching target location]
{\includegraphics[width=0.3\linewidth]{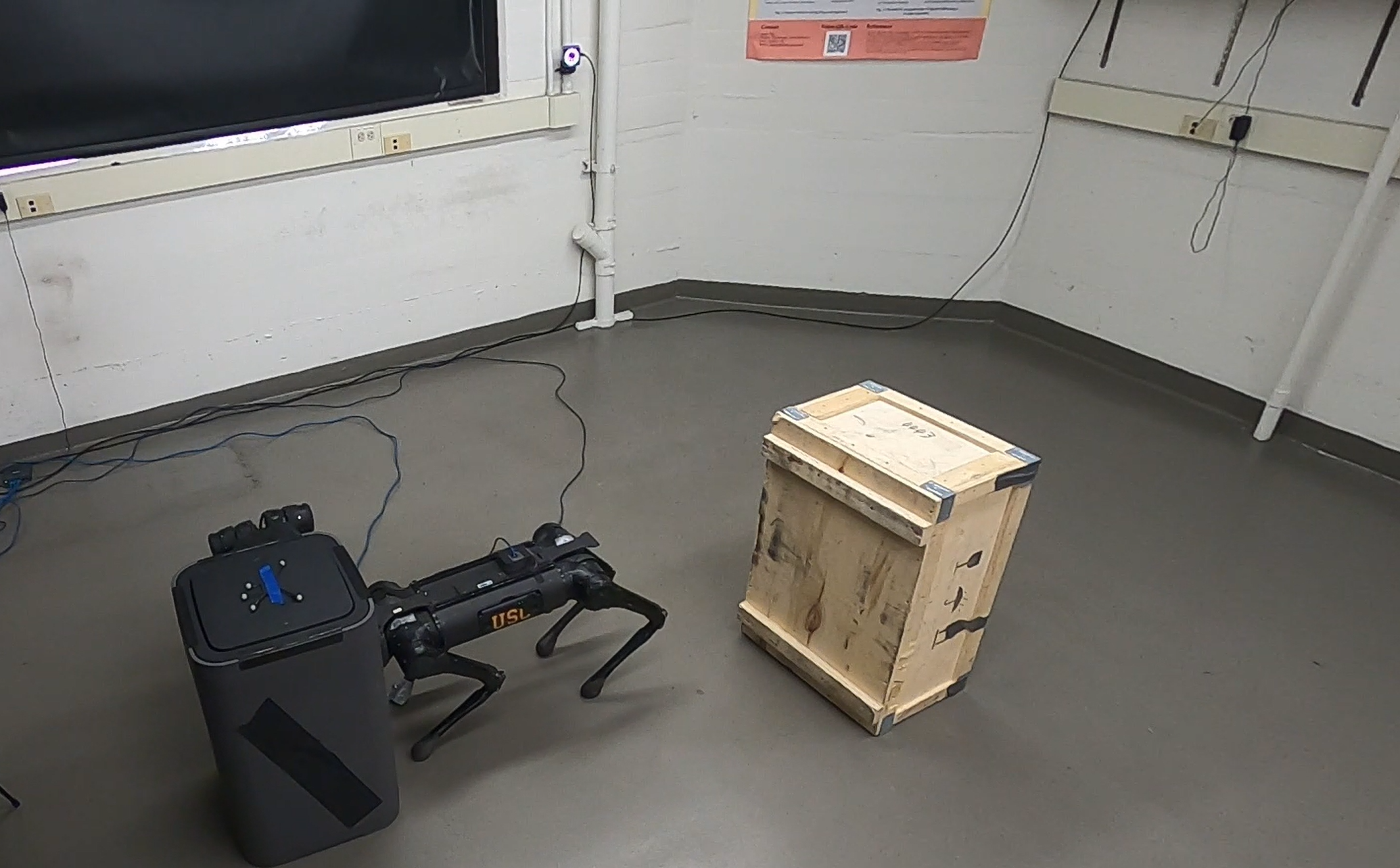}}
\caption{\textbf{Experimental Results.} Two Unitree robots (an Aliengo and an A1) manipulate the object while avoiding the obstacle.}
\label{fig: experiment}
\end{figure*}
\begin{figure}[t!]
\centering
\includegraphics[width=1\linewidth]{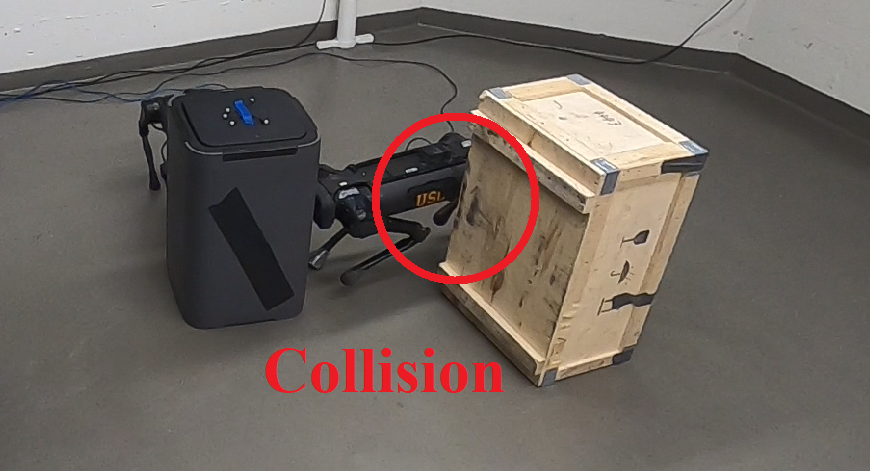}
\caption{\textbf{Experimental Result without Employing the CBFs Constraint for robot-obstacle collision.} The Aliengo robot collides with the obstacle while it wants to go around the obstacle.}
\label{fig: experiment without CBF}
\end{figure}
\subsubsection{Impact of Initial Robots Configuration}:
As previously mentioned, robots are not rigidly connected to the object in our manipulation task scenario. While this setup allows for some flexibility in adjusting each robot's position, especially to avoid collisions, it also introduces challenges for manipulation tasks. Without a rigid connection, applying force and torque is more difficult compared to if the robots were directly connected to the object. Each robot can only exert a force perpendicular to the object's surface and adjust its lateral positions to apply the proper torque and maneuver the object. These considerations are incorporated into the dynamic equations. Therefore, the initial configuration of robots around the object can influence the optimized trajectory. The trajectory is optimized to avoid making it overly difficult for the team of robots to manipulate the object. The planner achieves this through the cost associated with input minimization ($\|\bm{u}(t)\|_{\bm{R}u}$) and minimizing the change in the contact point location ($\|\bm{d}(t)- \bm{d}^*_{\text{prev}}\|_{\bm{Q}_d}$).

To illustrate the initial configuration effect, we conducted two simulations. In both simulations, the object's initial position is at the origin, and the target location is $\bm{x}_p^f = (5m,5m)$. Two static cubic obstacles were placed at coordinates $(3,2)$ and $(2,3)$; therefore, everything is symmetric for object manipulation purposes. Then, we use two different initial configurations for the team of robots, as shown in \figref{fig: from left} and \figref{fig: from right}. These configurations resulted in different optimized trajectories. In one configuration, the team of robots manipulated the object from the left side of the obstacles (\figref{fig: from left middle}), while in the other configuration, they navigated through the right side of the obstacles (\figref{fig: from right middle}). These results demonstrate how the initial configuration of the robot team can lead to different optimized trajectories to achieve optimal behavior.

\subsection{Hardware Experiments}

We also validated our approach using a team of two robots, a Unitree A1 and an Aliengo, to manipulate an object. In our setup, an obstacle blocks the path, preventing a direct trajectory to the target position. The object's weight is 6 kg (the box-only weight), with an additional unknown 3 kg load inside the box. Moreover, the dynamic equation lacks a friction model, introducing model uncertainty that the adaptive controller manages. The box's state is tracked using a motion capture system, and the obstacle's position is predetermined in the planner. The results demonstrated in the supplemental video and shown in snapshots in \figref{fig: experiment}.  

In this highly constrained setup, with a distance of approximately 3 meters between the start and target points and an obstacle with a surface area of nearly 0.5 square meters in between, the importance of Control Barrier Functions (CBFs) for preventing robot-obstacle collisions is crucial. To highlight this in our hardware experiment, we removed the CBF constraints for robot-obstacle collision in the motion planner and conducted the experiment. The results showed that while the object's trajectory remained safe, the Aliengo robot collided with the obstacle. This is illustrated in the snapshot provided in \figref{fig: experiment without CBF}.

\subsection{Handling Dynamic Obstacles}
\begin{figure*}[!t]
\centering
\subfloat[Initial Stage]{\includegraphics[angle=-90,width=0.3\linewidth]{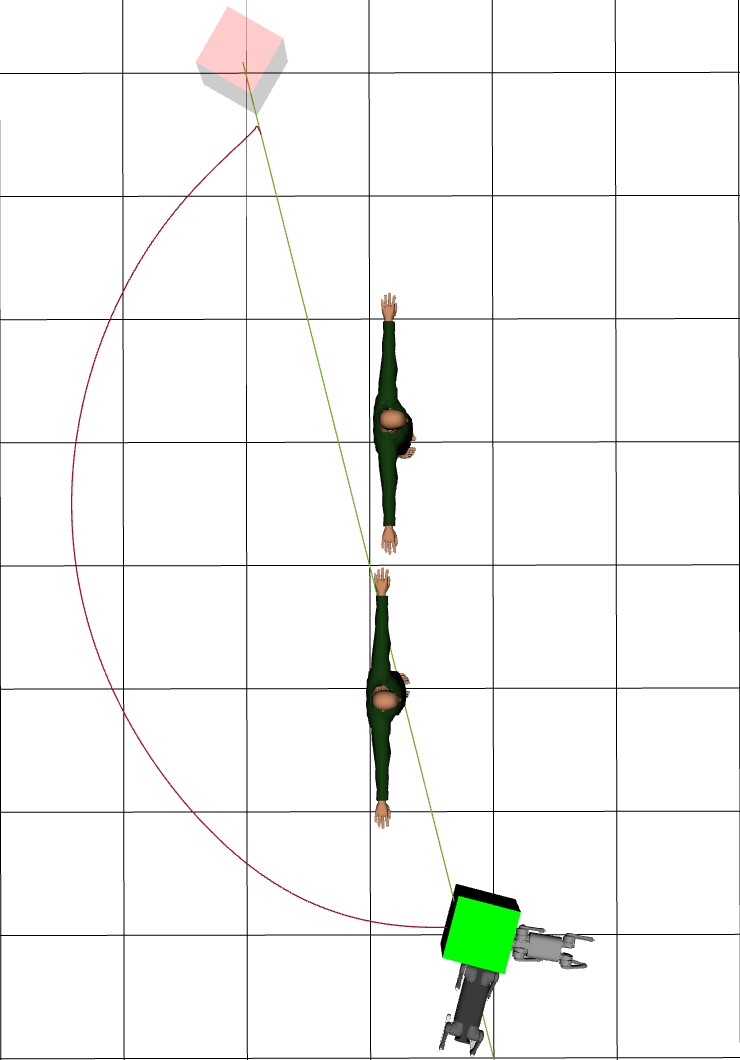}}
\hfill
\subfloat[Intermediate Stage]{\includegraphics[angle=-90,width=0.3\linewidth]{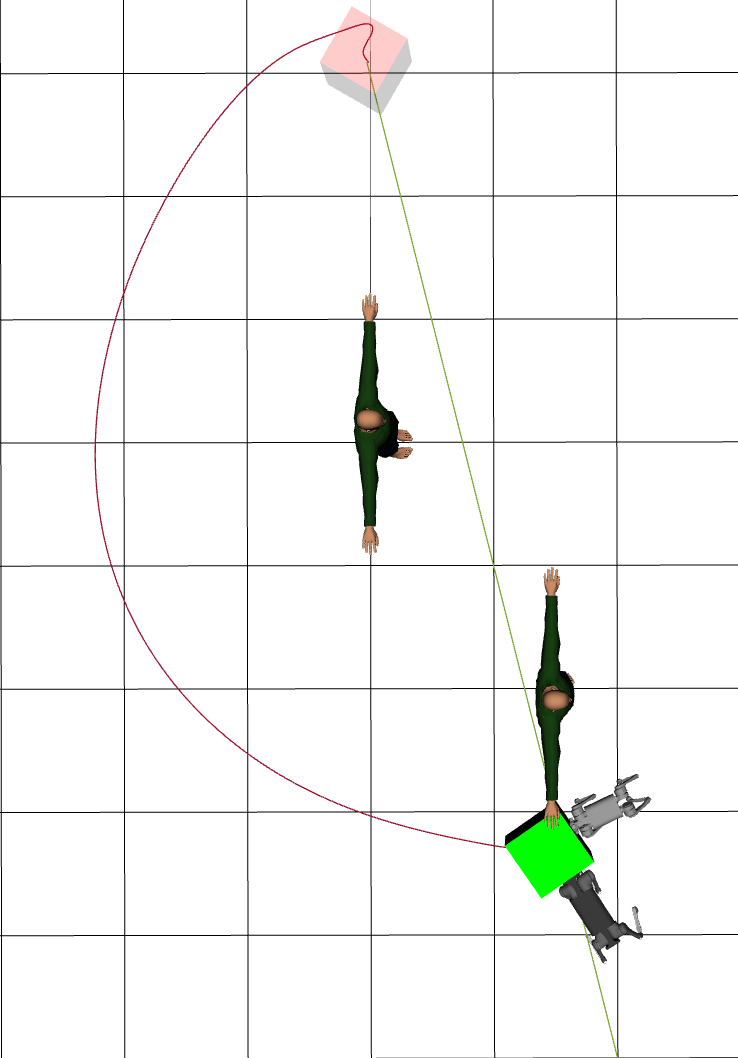}}
\hfill
\subfloat[Final Stage]{\includegraphics[angle=-90,width=0.3\linewidth]{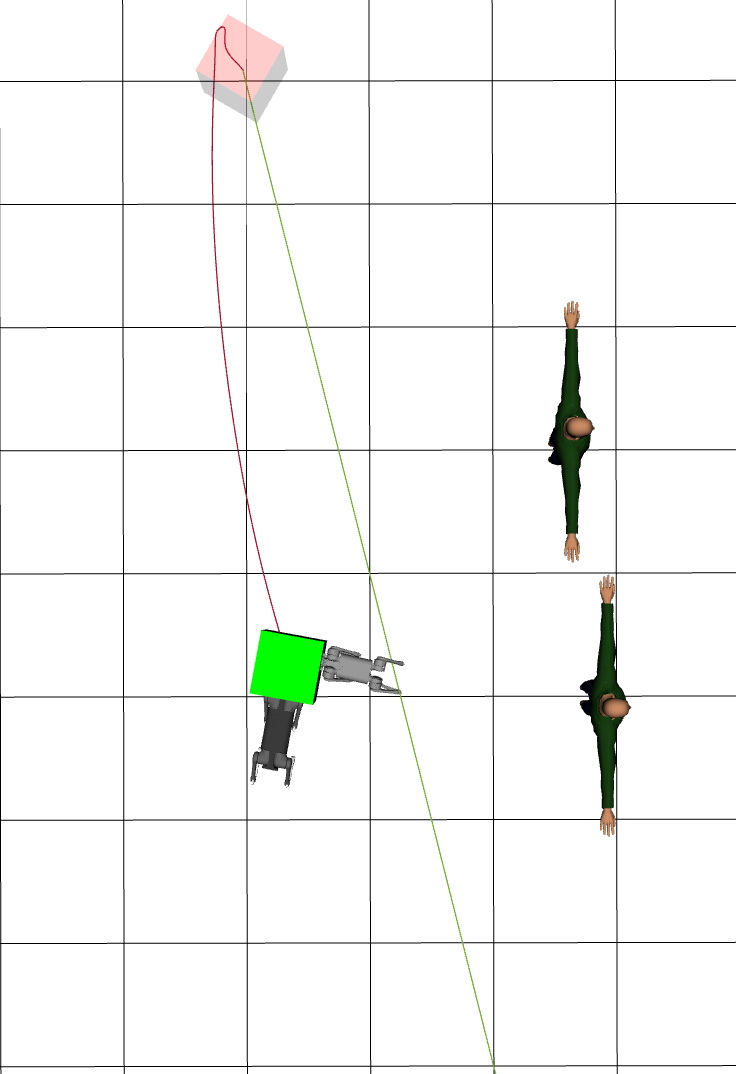}}
\caption{\textbf{Simulation of quadruped robots encountering dynamic obstacles.} The motion planner continuously optimizes the trajectory to avoid moving obstacles and ensure safety.}
\label{fig: dynamic_obstacles}
\end{figure*}

As previously mentioned, most practical applications involve quadruped robots operating in environments with moving elements. For instance, in warehouses where people are constantly working, ensuring safety is essential for robots operating in proximity to humans. In the following simulation, a team of robots encounters moving obstacles represented by humans in the environment. This simulation uses one Aliengo and one Go1 robot from Unitree, demonstrating that our approach can be implemented with different robot types and is not limited to a specific model. \figref{fig: dynamic_obstacles} and supplemental video show that the motion planner updates the trajectory when a human approaches the current optimized path, deviating to maintain a safe distance from the obstacle. This simulation illustrates how our proposed motion planner can use CBFs to manage collision avoidance with dynamic obstacles in the environment.

\subsection{Limitations}
A key limitation in our problem formulation is that the agents cannot directly apply a moment to the object. As a result, rotational motion is inherently coupled with translational motion since both are driven by the unidirectional forces applied by the robots to the object. These interaction constraints are embedded within the dynamic equations in the motion planner's MPC framework, as defined in \eqref{eq: interaction constraint}. This constraint, along with the velocity constraint \eqref{eq: vel_max}, which is restricted by the robots' speed limits, prevents the system's rapid maneuvering and rotations. Therefore, if the planner receives a sharp command, it often fails due to an infeasible optimization problem—particularly because the interaction constraint, enforced as a hard constraint, significantly limits the optimization feasibility. 
\section{Conclusion} \label{sec: conclusion}
In conclusion, this paper presents a safety-critical framework for object manipulation using a team of quadruped robots. The motion planner is designed in an MPC fashion, considering both stability and safety criteria to ensure the safe navigation of the object through an environment with obstacles. Additionally, the planner employs an adaptive controller to compensate for model uncertainties, addressing practical scenarios. Thus, the motion planner can handle objects with unknown mass and friction coefficients. By defining appropriate CBFs, the motion planner guarantees safety while calculating the optimized trajectory for the object. The manipulation task is then performed using a decentralized loco-manipulation controller for each agent, utilizing the optimized force and contact point locations provided by the motion planner.

We hope this work inspires future research in multi-entity-legged robot systems, which are crucial for real-world applications. In the future, we plan to overcome the limitations of the interaction constraint by developing a more convex dynamic formulation for the motion planner MPC, enabling greater maneuverability.


\bibliographystyle{SageH}
\bibliography{references}

\begin{thebibliography}{43}
\providecommand{\natexlab}[1]{#1}
\providecommand{\url}[1]{\texttt{#1}}
\providecommand{\urlprefix}{URL }
\expandafter\ifx\csname urlstyle\endcsname\relax
  \providecommand{\doi}[1]{DOI:\discretionary{}{}{}#1}\else
  \providecommand{\doi}{DOI:\discretionary{}{}{}\begingroup \urlstyle{rm}\Url}\fi

\bibitem[{Ames et~al.(2019)Ames, Coogan, Egerstedt, Notomista, Sreenath and Tabuada}]{Ames2019ControlApplications}
Ames AD, Coogan S, Egerstedt M, Notomista G, Sreenath K and Tabuada P (2019) {Control barrier functions: Theory and applications}.
\newblock In: \emph{2019 18th European Control Conference, ECC 2019}. IEEE.
\newblock ISBN 9783907144008, pp. 3420--3431.
\newblock \doi{10.23919/ECC.2019.8796030}.

\bibitem[{Ames et~al.(2014)Ames, Grizzle and Tabuada}]{Ames2014ControlControl}
Ames AD, Grizzle JW and Tabuada P (2014) {Control barrier function based quadratic programs with application to adaptive cruise control}.
\newblock \emph{Proceedings of the IEEE Conference on Decision and Control} 2015-February(February): 6271--6278.
\newblock \doi{10.1109/CDC.2014.7040372}.

\bibitem[{Ames et~al.(2017)Ames, Xu, Grizzle and Tabuada}]{Ames2017ControlSystems}
Ames AD, Xu X, Grizzle JW and Tabuada P (2017) {Control Barrier Function Based Quadratic Programs for Safety Critical Systems}.
\newblock \emph{IEEE Transactions on Automatic Control} 62(8): 3861--3876.
\newblock \doi{10.1109/TAC.2016.2638961}.

\bibitem[{Bledt et~al.(2018)Bledt, Powell, Katz, Di~Carlo, Wensing and Kim}]{Bledt2018}
Bledt G, Powell MJ, Katz B, Di~Carlo J, Wensing PM and Kim S (2018) {MIT Cheetah 3: Design and Control of a Robust, Dynamic Quadruped Robot}.
\newblock In: \emph{IEEE International Conference on Intelligent Robots and Systems}. Institute of Electrical and Electronics Engineers Inc.
\newblock ISBN 9781538680940, pp. 2245--2252.
\newblock \doi{10.1109/IROS.2018.8593885}.

\bibitem[{Chiu et~al.(2022)Chiu, Sleiman, Mittal, Farshidian and Hutter}]{Chiu2022}
Chiu JR, Sleiman JP, Mittal M, Farshidian F and Hutter M (2022) {A Collision-Free MPC for Whole-Body Dynamic Locomotion and Manipulation}.
\newblock In: \emph{Proceedings - IEEE International Conference on Robotics and Automation}. IEEE.
\newblock ISBN 9781728196817, pp. 4686--4693.
\newblock \doi{10.1109/ICRA46639.2022.9812280}.

\bibitem[{Culbertson et~al.(2021)Culbertson, Slotine and Schwager}]{Culbertson2021}
Culbertson P, Slotine JJ and Schwager M (2021) {Decentralized Adaptive Control for Collaborative Manipulation of Rigid Bodies; Decentralized Adaptive Control for Collaborative Manipulation of Rigid Bodies}.
\newblock \emph{IEEE Transactions on Robotics} 37(6).
\newblock \doi{10.1109/TRO.2021}.

\bibitem[{Di~Carlo et~al.(2018)Di~Carlo, Wensing, Katz, Bledt and Kim}]{DiCarlo2018}
Di~Carlo J, Wensing PM, Katz B, Bledt G and Kim S (2018) {Dynamic Locomotion in the MIT Cheetah 3 Through Convex Model-Predictive Control}.
\newblock In: \emph{IEEE International Conference on Intelligent Robots and Systems}. IEEE.
\newblock ISBN 9781538680940, pp. 7440--7447.
\newblock \doi{10.1109/IROS.2018.8594448}.

\bibitem[{Fadali and Visioli(2012)}]{Fadali2012}
Fadali MS and Visioli A (2012) \emph{{Digital Control Engineering: Analysis and Design}}.
\newblock Elsevier Science.
\newblock ISBN 9780123983244.

\bibitem[{Farshidian and {others}(2017)}]{Farshidian2017OCS2:Systems}
Farshidian F and {others} (2017) {OCS2: An open source library for Optimal Control of Switched Systems}.
\newblock \urlprefix\url{https://github.com/leggedrobotics/ocs2}.

\bibitem[{Feller and Ebenbauer(2017)}]{Feller2017AFunctions}
Feller C and Ebenbauer C (2017) {A stabilizing iteration scheme for model predictive control based on relaxed barrier functions}.
\newblock \emph{Automatica} 80: 328--339.
\newblock \doi{10.1016/j.automatica.2017.02.001}.

\bibitem[{Ferreau et~al.(2014)Ferreau, Kirches, Potschka, Bock and Diehl}]{Ferreau2014QpOASES:Programming}
Ferreau HJ, Kirches C, Potschka A, Bock HG and Diehl M (2014) {qpOASES: a parametric active-set algorithm for quadratic programming}.
\newblock \emph{Mathematical Programming Computation} 6(4): 327--363.
\newblock \doi{10.1007/S12532-014-0071-1/TABLES/5}.

\bibitem[{Ferrolho et~al.(2023)Ferrolho, Ivan, Merkt, Havoutis and Vijayakumar}]{Ferrolho2023RoLoMa:Arms}
Ferrolho H, Ivan V, Merkt W, Havoutis I and Vijayakumar S (2023) {RoLoMa: robust loco-manipulation for quadruped robots with arms}.
\newblock \emph{Autonomous Robots} 47(8): 1463--1481.
\newblock \doi{10.1007/s10514-023-10146-0}.

\bibitem[{Fink et~al.(2008)Fink, Ani~Hsieh and Kumar}]{Fink2008}
Fink J, Ani~Hsieh M and Kumar V (2008) {Multi-robot manipulation via caging in environments with obstacles}.
\newblock In: \emph{Proceedings - IEEE International Conference on Robotics and Automation}. IEEE.
\newblock ISBN 9781424416479, pp. 1471--1476.
\newblock \doi{10.1109/ROBOT.2008.4543409}.

\bibitem[{Focchi et~al.(2017)Focchi, del Prete, Havoutis, Featherstone, Caldwell and Semini}]{Focchi2017}
Focchi M, del Prete A, Havoutis I, Featherstone R, Caldwell DG and Semini C (2017) {High-slope terrain locomotion for torque-controlled quadruped robots}.
\newblock \emph{Autonomous Robots} 41(1): 259--272.
\newblock \doi{10.1007/s10514-016-9573-1}.

\bibitem[{Grandia et~al.(2023)Grandia, Jenelten, Yang, Farshidian and Hutter}]{Grandia2022a}
Grandia R, Jenelten F, Yang S, Farshidian F and Hutter M (2023) {Perceptive Locomotion Through Nonlinear Model-Predictive Control}.
\newblock \emph{IEEE Transactions on Robotics} 39(5): 3402--3421.
\newblock \doi{10.1109/TRO.2023.3275384}.

\bibitem[{Hauser and Saccon(2006)}]{Hauser2006AConstraints}
Hauser J and Saccon A (2006) {A Barrier Function Method for the Optimization of Trajectory Functionals with Constraints}.
\newblock In: \emph{Proceedings of the 45th IEEE Conference on Decision and Control}. IEEE.
\newblock ISBN 1-4244-0171-2, pp. 864--869.
\newblock \doi{10.1109/CDC.2006.377331}.

\bibitem[{Hu et~al.(1995)Hu, Goldenberg and Zhou}]{Hu1995}
Hu Yr, Goldenberg AA and Zhou C (1995) {Motion and Force Control of Coordinated Robots During Constrained Motion Tasks}.
\newblock \emph{The International Journal of Robotics Research} 14(4): 351--365.
\newblock \doi{10.1177/027836499501400404}.

\bibitem[{Khatib(1988)}]{Khatib1988ObjectSystem}
Khatib O (1988) {Object manipulation in a multi-effector robot system}.
\newblock In: \emph{Proceedings of the 4th International Symposium on Robotics Research}. Cambridge, MA, USA: MIT Press.
\newblock ISBN 0262022729, pp. 137--144.

\bibitem[{Khatib et~al.(1999)Khatib, Yokoi, Brock, Chang and Casal}]{Khatib1999RobotsCapabilities}
Khatib O, Yokoi K, Brock O, Chang K and Casal A (1999) {Robots in Human Environments: Basic Autonomous Capabilities}.
\newblock \emph{The International Journal of Robotics Research} 18(7): 684--696.
\newblock \doi{10.1177/02783649922066501}.

\bibitem[{Khatib et~al.(1996)Khatib, Yokoi, Chang, Ruspini, Holmberg and Casal}]{Khatib1996}
Khatib O, Yokoi K, Chang K, Ruspini D, Holmberg R and Casal A (1996) {Coordination and decentralized cooperation of multiple mobile manipulators}.
\newblock \emph{Journal of Robotic Systems} 13(11): 755--764.
\newblock \doi{10.1002/(SICI)1097-4563(199611)13:11<755::AID-ROB6>3.0.CO;2-U}.

\bibitem[{Kim et~al.(2023)Kim, Fawcett, Kamidi, Ames and Hamed}]{Kim2023LayeredApproaches}
Kim J, Fawcett RT, Kamidi VR, Ames AD and Hamed KA (2023) {Layered Control for Cooperative Locomotion of Two Quadrupedal Robots: Centralized and Distributed Approaches}.
\newblock \emph{IEEE Transactions on Robotics} 39(6): 4728--4748.
\newblock \doi{10.1109/TRO.2023.3319896}.

\bibitem[{Li and Nguyen(2021)}]{Li2021}
Li J and Nguyen Q (2021) {Force-and-moment-based Model Predictive Control for Achieving Highly Dynamic Locomotion on Bipedal Robots}.
\newblock In: \emph{Proceedings of the IEEE Conference on Decision and Control}, volume 2021-Decem. IEEE.
\newblock ISBN 9781665436595, pp. 1024--1030.
\newblock \doi{10.1109/CDC45484.2021.9683500}.

\bibitem[{Li et~al.(2008)Li, Ge and Wang}]{Li2008}
Li Z, Ge SS and Wang Z (2008) {Robust adaptive control of coordinated multiple mobile manipulators}.
\newblock \emph{Mechatronics} 18(5-6): 239--250.
\newblock \doi{10.1016/j.mechatronics.2008.01.001}.

\bibitem[{Liu and Arimoto(1998)}]{Liu1998DecentralizedCooperations}
Liu YH and Arimoto S (1998) {Decentralized Adaptive and Nonadaptive Position/Force Controllers for Redundant Manipulators in Cooperations}.
\newblock \emph{The International Journal of Robotics Research} 17(3): 232--247.
\newblock \doi{10.1177/027836499801700302}.

\bibitem[{Nguyen and Sreenath(2016)}]{Nguyen2016ExponentialConstraints}
Nguyen Q and Sreenath K (2016) {Exponential Control Barrier Functions for enforcing high relative-degree safety-critical constraints}.
\newblock In: \emph{Proceedings of the American Control Conference}, volume 2016-July. IEEE.
\newblock ISBN 9781467386821, pp. 322--328.
\newblock \doi{10.1109/ACC.2016.7524935}.

\bibitem[{Nocedal and Wright(2006)}]{Nocedal2006NumericalOptimization}
Nocedal J and Wright SJ (2006) \emph{{Numerical Optimization}}.
\newblock Springer Series in Operations Research and Financial Engineering. Springer New York.
\newblock ISBN 978-0-387-40065-5.
\newblock \doi{10.1007/978-0-387-40065-5}.

\bibitem[{Pandit et~al.(2024)Pandit, Gupta, Gadde, Johnson, Shrestha, Duan, Dao and Fern}]{Pandit2024LearningTransport}
Pandit B, Gupta A, Gadde MS, Johnson A, Shrestha AK, Duan H, Dao J and Fern A (2024) {Learning Decentralized Multi-Biped Control for Payload Transport}.
\newblock \urlprefix\url{https://arxiv.org/abs/2406.17279v1}.

\bibitem[{Prattichizzo and Trinkle(2008)}]{Prattichizzo2008}
Prattichizzo D and Trinkle JC (2008) {Grasping}.
\newblock In: \emph{Springer Handbook of Robotics}. Berlin, Heidelberg: Springer Berlin Heidelberg, pp. 671--700.
\newblock \doi{10.1007/978-3-540-30301-5{\_}29}.

\bibitem[{Rigo et~al.(2023)Rigo, Chen, Gupta and Nguyen}]{Rigo2023}
Rigo A, Chen Y, Gupta SK and Nguyen Q (2023) {Contact Optimization for Non-Prehensile Loco-Manipulation via Hierarchical Model Predictive Control}.
\newblock In: \emph{Proceedings - IEEE International Conference on Robotics and Automation}, volume 2023-May.
\newblock ISBN 9798350323658, pp. 9945--9951.
\newblock \doi{10.1109/ICRA48891.2023.10160507}.

\bibitem[{Sleiman et~al.(2021)Sleiman, Farshidian, Minniti and Hutter}]{Sleiman2021}
Sleiman JP, Farshidian F, Minniti MV and Hutter M (2021) {A Unified MPC Framework for Whole-Body Dynamic Locomotion and Manipulation}.
\newblock \emph{IEEE Robotics and Automation Letters} 6(3): 4688--4695.
\newblock \doi{10.1109/LRA.2021.3068908}.

\bibitem[{Slotine and Li(1991)}]{Slotine1991}
Slotine JJE and Li W (1991) \emph{{Applied nonlinear control}}, volume 199.
\newblock Prentice hall Englewood Cliffs, NJ.

\bibitem[{Sombolestan et~al.(2021)Sombolestan, Chen and Nguyen}]{Sombolestan2021}
Sombolestan M, Chen Y and Nguyen Q (2021) {Adaptive Force-based Control for Legged Robots}.
\newblock In: \emph{IEEE International Conference on Intelligent Robots and Systems}. IEEE.
\newblock ISBN 9781665417143, pp. 7440--7447.
\newblock \doi{10.1109/IROS51168.2021.9636393}.

\bibitem[{Sombolestan and Nguyen(2023{\natexlab{a}})}]{Sombolestan2023}
Sombolestan M and Nguyen Q (2023{\natexlab{a}}) {Hierarchical Adaptive Control for Collaborative Manipulation of a Rigid Object by Quadrupedal Robots}.
\newblock In: \emph{2023 IEEE/RSJ International Conference on Intelligent Robots and Systems (IROS)}. IEEE.
\newblock ISBN 978-1-6654-9190-7, pp. 2752--2759.
\newblock \doi{10.1109/IROS55552.2023.10341700}.

\bibitem[{Sombolestan and Nguyen(2023{\natexlab{b}})}]{Sombolestan2023b}
Sombolestan M and Nguyen Q (2023{\natexlab{b}}) {Hierarchical Adaptive Loco-manipulation Control for Quadruped Robots}.
\newblock In: \emph{Proceedings - IEEE International Conference on Robotics and Automation}, volume 2023-May. IEEE.
\newblock ISBN 9798350323658, pp. 12156--12162.
\newblock \doi{10.1109/ICRA48891.2023.10160523}.

\bibitem[{Sombolestan and Nguyen(2024)}]{Sombolestan2024}
Sombolestan M and Nguyen Q (2024) {Adaptive-Force-Based Control of Dynamic Legged Locomotion Over Uneven Terrain}.
\newblock \emph{IEEE Transactions on Robotics} 40: 2462--2477.
\newblock \doi{10.1109/TRO.2024.3381554}.

\bibitem[{Sontag(1999)}]{Sontag1999}
Sontag ED (1999) {Control-Lyapunov functions}.
\newblock In: Blondel V, Sontag ED, Vidyasagar M and Willems JC (eds.) \emph{Open Problems in Mathematical Systems and Control Theory}. London: Springer London.
\newblock ISBN 978-1-4471-0807-8, pp. 211--216.
\newblock \doi{10.1007/978-1-4471-0807-8{\_}40}.

\bibitem[{Tallamraju et~al.(2019)Tallamraju, Verma, Sripada, Agrawal and Karlapalem}]{Tallamraju2019EnergyValidation}
Tallamraju R, Verma P, Sripada V, Agrawal S and Karlapalem K (2019) {Energy Conscious Over-actuated Multi-Agent Payload Transport Robot: Simulations and Preliminary Physical Validation}.
\newblock In: \emph{2019 28th IEEE International Conference on Robot and Human Interactive Communication (RO-MAN)}. IEEE.
\newblock ISBN 978-1-7281-2622-7, pp. 1--7.
\newblock \doi{10.1109/RO-MAN46459.2019.8956442}.

\bibitem[{Tarn et~al.(1986)Tarn, Bejczy and Yun}]{Tarn1986COORDINATEDARMS}
Tarn T, Bejczy A and Yun X (1986) {Coordinated control of two robot arms}.
\newblock In: \emph{Proceedings. 1986 IEEE International Conference on Robotics and Automation}. IEEE.
\newblock ISBN 0818606959, pp. 1193--1202.
\newblock \doi{10.1109/ROBOT.1986.1087606}.

\bibitem[{Turrisi et~al.(2024)Turrisi, Schulze, Medeiros, Semini and Barasuol}]{Turrisi2024PACC:Control}
Turrisi G, Schulze L, Medeiros VS, Semini C and Barasuol V (2024) {PACC: A Passive-Arm Approach for High-Payload Collaborative Carrying with Quadruped Robots Using Model Predictive Control}.
\newblock \urlprefix\url{https://arxiv.org/abs/2403.19862v2}.

\bibitem[{Verginis et~al.(2017)Verginis, Mastellaro and Dimarogonas}]{Verginis2017}
Verginis CK, Mastellaro M and Dimarogonas DV (2017) {Robust Quaternion-based Cooperative Manipulation without Force/Torque Information}.
\newblock \emph{IFAC-PapersOnLine} 50(1): 1754--1759.
\newblock \doi{10.1016/j.ifacol.2017.08.526}.

\bibitem[{Wolfslag et~al.(2020)Wolfslag, McGreavy, Xin, Tiseo, Vijayakumar and Li}]{Wolfslag2020OptimisationRobots}
Wolfslag WJ, McGreavy C, Xin G, Tiseo C, Vijayakumar S and Li Z (2020) {Optimisation of Body-ground Contact for Augmenting the Whole-Body Loco-manipulation of Quadruped Robots}.
\newblock In: \emph{2020 IEEE/RSJ International Conference on Intelligent Robots and Systems (IROS)}. IEEE.
\newblock ISBN 978-1-7281-6212-6, pp. 3694--3701.
\newblock \doi{10.1109/IROS45743.2020.9341498}.

\bibitem[{Yang et~al.(2022)Yang, Sue, Li, Yang, Shen, Chi, Rai, Zeng and Sreenath}]{Yang2022b}
Yang C, Sue GN, Li Z, Yang L, Shen H, Chi Y, Rai A, Zeng J and Sreenath K (2022) {Collaborative Navigation and Manipulation of a Cable-Towed Load by Multiple Quadrupedal Robots}.
\newblock \emph{IEEE Robotics and Automation Letters} 7(4): 10041--10048.
\newblock \doi{10.1109/LRA.2022.3191170}.

\bibitem[{Zimmermann et~al.(2021)Zimmermann, Poranne and Coros}]{Zimmermann2021}
Zimmermann S, Poranne R and Coros S (2021) {Go Fetch! - Dynamic Grasps using Boston Dynamics Spot with External Robotic Arm}.
\newblock In: \emph{Proceedings - IEEE International Conference on Robotics and Automation}, volume 2021-May. IEEE.
\newblock ISBN 9781728190778, pp. 1170--1176.
\newblock \doi{10.1109/ICRA48506.2021.9561835}.

\end{thebibliography}

\end{document}